\documentclass[conference]{IEEEtran}
\usepackage{times}

\usepackage[numbers]{natbib}
\usepackage{multicol}
\usepackage[bookmarks=true]{hyperref}
\usepackage{xcolor}
\usepackage{amsmath}
\usepackage{graphicx} 
\usepackage{subcaption}
\usepackage{makecell}
\usepackage{rotating}
\usepackage{booktabs} 
\usepackage{array}    
\usepackage{float}

\usepackage{amssymb} 

\newcommand{\benchmark}{The Colosseum}

\newcommand{\myrotcell}[1]{\rotcell{\makebox[0pt][l]{#1}}}

\newcommand{\benchname}[0]{\textsc{The Colosseum}}
\newcommand{\variationfactors}[0]{\texttt{perturbation\_factors}}
\newcommand{\numvariationfactors}[0]{14}
\newcommand{\numsotamodels}[0]{5}
\newcommand{\numcolors}[0]{20}
\newcommand{\numtextures}[0]{213}
\newcommand{\numdistractors}[0]{78}

\begin{document}

\title{\includegraphics[height=7mm]{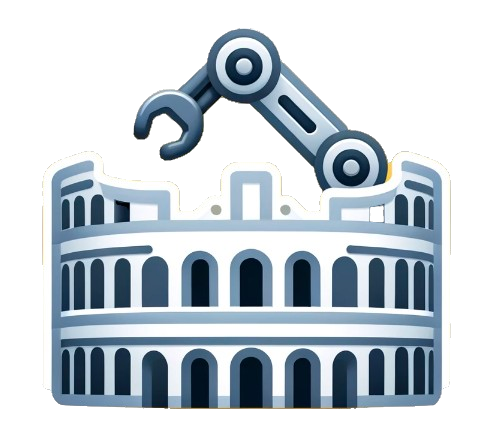}\benchname:
A Benchmark for Evaluating Generalization for Robotic Manipulation
}

\author{Author Names Omitted for Anonymous Review. Paper-ID 41}



%

\author{
\authorblockN{Wilbert Pumacay\authorrefmark{1}
\authorblockA{Universidad Catolica San Pablo}\\}

\authorblockN{Ranjay Krishna
\authorblockA{University of Washington}
\authorblockA{Allen Institute for Artifical Intelligence}}
\and

\authorblockN{Ishika Singh\authorrefmark{1}
\authorblockA{University of Southern California}\\}

\authorblockN{Jesse Thomason
\authorblockA{University of Southern California}\\}
\authorblockA{\authorrefmark{1}equal contribution}
\authorblockA{\href{https://robot-colosseum.github.io/}{robot-colosseum.github.io}}
\and

\authorblockN{Jiafei Duan\authorrefmark{1}
\authorblockA{University of Washington}\\}

\authorblockN{Dieter Fox
\authorblockA{University of Washington}
\authorblockA{NVIDIA}}
}

\maketitle

\begin{abstract}
To realize effective large-scale, real-world robotic applications, we must evaluate how well our robot policies adapt to changes in environmental conditions. Unfortunately, a majority of studies evaluate robot performance in environments closely resembling or even identical to the training setup. 
We present \benchname, a novel simulation benchmark, with 20 diverse manipulation tasks, that enables systematical evaluation of models across \numvariationfactors\ axes of environmental perturbations.
These perturbations include changes in color, texture, and size of objects, table-tops, and backgrounds;  we also vary lighting, distractors, physical properties perturbations and camera pose.
Using \benchname, we compare 5 state-of-the-art manipulation models to reveal that their success rate degrades between 30-50\% across these perturbation factors.
When multiple perturbations are applied in unison, the success rate degrades $\geq$75\%.
We identify that changing the number of distractor objects, target object color, or lighting conditions are the perturbations that reduce model performance the most.
To verify the ecological validity of our results, we show that our results in simulation are  correlated ($\bar{R}^2 = 0.614$) to similar perturbations in real-world experiments.
We open source code for others to use \benchname, and also release code to 3D print the objects used to replicate the real-world perturbations.
Ultimately, we hope that \benchname\ will serve as a benchmark to identify modeling decisions that systematically improve generalization for manipulation.
\end{abstract}

\IEEEpeerreviewmaketitle

\section{Introduction}
\label{sec:intro}

\begin{figure}[t]
  \centering
  \includegraphics[width=\linewidth]{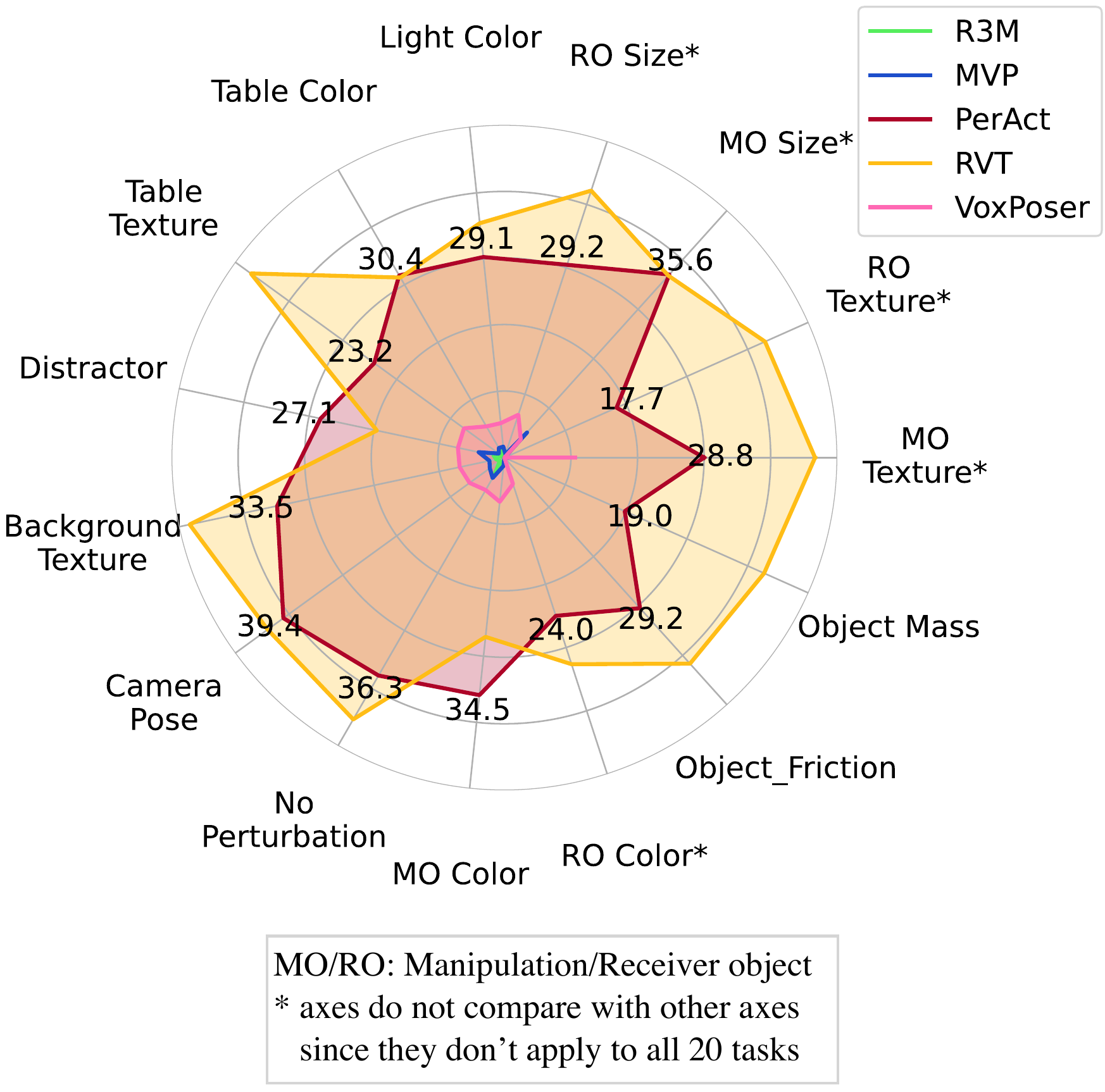}
  \caption{\textbf{Evaluating generalization with \benchname.} 
  Task-averaged success rate for 5 SotA robotic manipulation policies over \numvariationfactors\ perturbation factors and 20 robotic manipulation tasks.
  Changes in RGB input space affects all models due to end-to-end RGB-based training. 
  Image-based models are also affected by camera pose change, while models without in-the-wild pretraining suffer in the presence of distractors.}
  \label{fig:results}
\end{figure}

\begin{figure*}[t!]
  \centering
  \includegraphics[width=\linewidth]{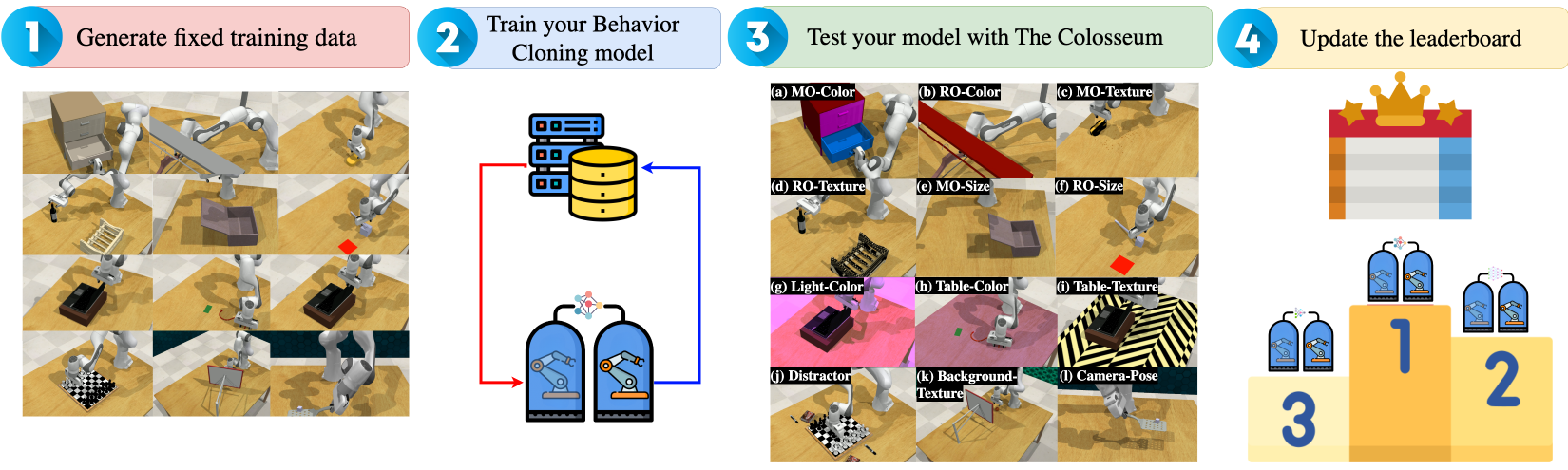}
  \caption{\textbf{\benchname\ Challenge.} This challenge is designed to enhance generalization of Behavior Cloning (BC) models in robotic manipulation tasks. It involves four key phases: 1) Participants generate a standard training dataset from 20 tasks with 100 demonstrations each, without \variationfactors. 2) Participants train their BC models using this standardized dataset. 3) The models are restricted to evaluate over a fixed 25 episodes across \numvariationfactors\ different \variationfactors. 4) Models are ranked on a leaderboard based on the percentage change in their performance across these factors. We've shown that simulation aligns with real-world evaluation, so participants can expect similar generalization when participating in the simulation benchmark.}
  \label{fig:challenge}
\end{figure*}

The promise of robotics requires ubiquity. For effective real-world deployment, robots must operate in a variety of environments. When asked to turn on a stove, a robot should be able to turn the stove's knob, regardless of the size of the knob, irrespective of the kitchen's backdrop, invariant to the kitchen counter's texture, during the day, or even under a dim evening light. 
Unfortunately, a majority of studies evaluate robot performance in environments closely resembling or even identical to the training setup~\cite{peract, rvt, RT1, chi2023diffusionpolicy}.

Naturally, generalization to environmental conditions has been a large focus in recent literature. 
Both Reinforcement Learning (RL)~\cite{tobin2017domain,nakamoto2023cal,mehta2020active} and Behavior Cloning (BC)~\cite{act-aloha,peract,r3m,rvt} struggle with generalization if not trained on sufficiently representative data. 
In response, robotics researchers have recently released large-scale diverse behavior cloning datasets, with trajectories collected either in simulation~\cite{duan2023ar2,ehsani2023imitating} or in the real world~\cite{rtx}.
With these datasets, different techniques—including data augmentation~\cite{laskin2020reinforcement,yarats2021mastering,peract}, pre-training on large vision and robot datasets for BC~\cite{r3m,mvp,rt2}, and incorporating 3D priors~\cite{peract,cliport,gervet2023act3d,sundaresan2023kite}—claim to improve generalization for manipulation tasks.
Although these techniques showcase improvements, the evaluation benchmarks are not designed to stress-test the policies against systematic perturbations to the environment. 
 
We introduce \benchname, a comprehensive benchmark aimed at systematically evaluating the generalization of robot manipulation to environmental perturbations. \benchname\ introduces perturbations across 20 different tasks from the RLBench~\cite{james2020rlbench} framework, spanning \numvariationfactors\ dimensions of perturbations. 
These perturbations include object color, object texture, object size, table color, table texture, the presence of distractor objects, changes to the camera pose, and changes to physical properties like friction and mass. 
\benchname\ also includes a parallel real world evaluation with task setups and objects reproducible via open-sourced 3D printing models.

We evaluate four state of the art BC models using \benchname\ and draw insights into answers for critical research questions on generalization for BC policies.
Considering 3D versus 2D reasoning methods, we find that 3D-based BC models demonstrate superiority over 2D-based methods in terms of overall task performance when using a fixed set of training data while also achieving better robustness to environmental perturbations.
Among 2D and 3D models, distractors, color-related and lighting perturbations have the most significant impact on task success.
Conversely, perturbations on object size had less impact in both settings.
Finally, we establish a strong correlation between falling task success under perturbations in simulations and those observed in real-world scenarios for the same tasks, suggesting that \benchname\ evaluations in simulation give reliable insight into real world generalization at a fraction of the setup cost.
\benchname\ challenge and leaderboard (Figure~\ref{fig:challenge}) will provide as a unified platform to develop, evaluate, and compare future robotic manipulation methods that stand the test of robustness and generalization.

\begin{table*}
\label{tab:table-compare}
\centering
\begin{tabular}{llp{2.5cm}p{2cm}p{2cm}p{2cm}p{2cm}}
\toprule
Benchmark & Simulator & No. of perturbations & No. of tasks & Physical perturbation & Real-world reproducibility & No. of SoTA models  \\
\midrule
GROOT~\cite{zhu2023learning} & LIBERO~\cite{liu2024libero} & 3 & 3 &$\times$  & $\checkmark$ & 4\\
VLMBench~\cite{zheng2022vlmbench} & RLBench~\cite{james2020rlbench} & 8 & 8 & $\times$ & $\times$ & 3\\
KitchenShift~\cite{xing2021kitchenshift} & Isaac Sim~\cite{makoviychuk2021isaac} & 7 & 3 & $\times$ & $\times$& 5\\
FactorWorld~\cite{Xie2023decomposing} & MuJoCo~\cite{todorov2012mujoco}& 11 & 19 & $\times$  & $\times$ & 2\\
\benchname\ (ours) & RLBench~\cite{james2020rlbench} & $14^{*}$ & 20 & $\checkmark$ & $\checkmark$ & 5\\

\bottomrule
\end{tabular}
\caption{\textbf{Generalization benchmarks comparison.} \benchname\ is the largest and most diverse benchmark for evaluating generalization in robotic manipulation policies, covering a wide range of variations and tasks. It is also the first to incorporate physical property perturbations. Similar to GROOT \cite{zhu2023learning}, we offer comprehensive instructions and 3D printed components for replicating real-world experiments. Additionally, we evaluated a variety of robot manipulation policies. One asterisk (*) indicates that, unlike the other benchmarks, the object's position is considered a default perturbation and was not counted as an additional perturbation.}
\end{table*}

\section{Related Work}
\label{sec:related}
Prior works have made contributions towards benchmarking robot manipulation, developing robust models, and demonstrating generalization.
\benchname\ builds on these efforts to create a systematic evaluation of multiple forms of test time generalization a trained policy may face.

\subsection{Robotic Manipulation Benchmarks}
Benchmarks in computer vision have significantly advanced the development of more generalized vision systems in recent decades by introducing numerous challenges and leaderboards \cite{deng2009imagenet,krishna2017visual, lin2014microsoft}, subsequently scaling into extensive foundational vision models \cite{chen2023minigpt, bai2023qwen, liu2023llava}. Similarly, robotics datasets have demonstrated considerable diversity across various dimensions \cite{dasari2019robonet,duan2020actionet,ebert2021bridge,kalashnikov2018scalable,yuan2024rl}, particularly with the evolution of imitation learning and, more specifically, behavior cloning (BC). This progress has led to a proliferation of datasets and benchmarks aimed at assessing BC model task performance. Furthermore, most of these robotic benchmarks focus on evaluating model's capability to adapt to new tasks by changing the nature of the task \cite{james2020rlbench,zhu2020robosuite}, it's functionalities\cite{heo2023furniturebench, luo2024fmb}, or even environment \cite{deitke2022️,puig2023habitat}. However, a gap remains in systematically evaluating and comparing different BC models on a large-scale, both in simulation and in real-world. Additionally, many of the benchmarks and datasets with perturbations have been specifically defined or curated as part of various BC or reinforcement learning works \cite{zhu2023learning,hansen2021softda,yuan2024rl}. However, they do not provide an unified framework to evaluate all potential perturbations for generalization, primarily because they are not the main focus of these works.

Factor World~\cite{Xie2023decomposing} and KitchenShift~\cite{xing2021kitchenshift} are similar efforts to \benchname. However, Factor World encompasses only 11 variation factors across 19 tasks, whereas KitchenShift contains 7 variation factors across 3 tasks. In contrast, The COLOSSEUM boasts \numvariationfactors\ factors of variation over 20 tasks.
Beyond that, \benchname\ supports both 2D and 3D models, unlike Factor World which only evaluates on 2D visual-motor policies. 
Furthermore, we employed 3D printed objects to test with a Franka Panda robot arm to enable easy replication of our real-world experiments.

\subsection{Robotic Manipulation Methods}

There are myriad approaches that model robotic manipulation in simulation and real world in various different ways.
Vanilla RL or BC~\cite{ng2006autonomous,mnih2013playing,schulman2017proximal,pomerleau1988alvinn,zhang2018deep} have been the popular choice since a long time, where a Multi-layer Perceptron (MLP)~\cite{popescu2009multilayer} or a Recurrent Neural Network (RNN)~\cite{rumelhart1986learning} type models use either low-dimensional object poses~\cite{sundermeyer2021contact} or images~\cite{chi2023diffusionpolicy} as the state input and predict continuous actions for the robot's controller in end-effector or joint space~\cite{shahid2020learning}.
An emerging area of work takes the path of representation learning, either by pretraining a model with external knowledge~\cite{r3m, mvp} or using pretrained representations from vision or language domains~\cite{cliport, huang2023voxposer, RT1, rt2}.
Recently, training generalist models on large-scale real robot datasets, collected across several robotics research labs~\cite{rtx}, to obtain a diverse set of skills in diverse environments and robots, have shown promising results in effectively scaling robust robot policies. 
Another line of work uses diffusion architecture to learn to generate robot trajectories in state or action space using the denoising process~\cite{condionalgen, chi2023diffusionpolicy, playfusion}.  
Some works have proposed distilling neural feature field representations for downstream BC~\cite{Ze2023GNFactor} or RL~\cite{NeRF-RL}. \citet{gervet2023act3d} learn with 3D feature field created from 2D pretrained image features and adaptive scene resolution to compute 3D action maps of high spatial resolution. 
\citet{wang2023mimicplay} and \citet{mandlekar2023mimicgen} propose scalable learning by watching humans or automatically generating large-scale datasets from a few human demonstrations.
Another work uses large-scale TAMP generated data in simulation to learn a scalable multitask policy~\cite{dalal2023optimus}.
Recent robotic manipulation works have also proposed learning keypoint action prediction~\cite{c2farm, peract, sundaresan2023kite, rvt} or action chucking~\cite{act-aloha} instead of predicting continuous control actions. We select a few recent SOTA methods, that have been applied to both simulation and real world, to evaluate on \benchname\ including 2D and 3D learning methods, that operate with keypoint action prediction.

\begin{figure}[t]
  \centering
  \includegraphics[width=\linewidth]{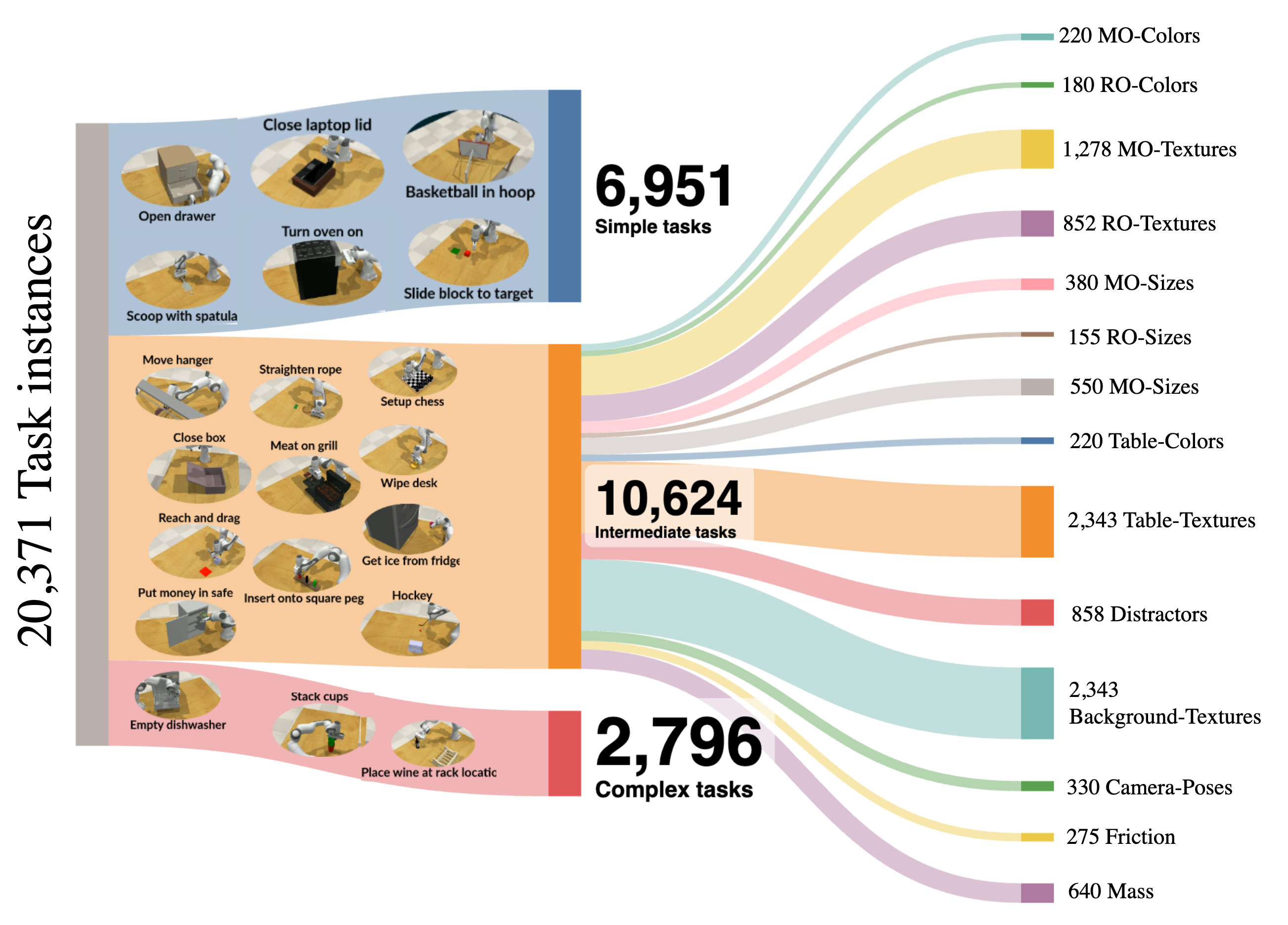}
  \caption{\textbf{\benchname\ benchmark distribution.} This benchmark encompasses \numvariationfactors\ \variationfactors\ within 20 distinct RLBench tasks, categorized into three tiers (simple, intermediate, and complex) according to the number of way-points involved (task horizon). Collectively, \benchname\ presents 20,371 unique task perturbations instances.}
  \label{fig:benchmark-data}
\end{figure}

\subsection{Generalization in Robotic Manipulation}
Traditionally, enhancing generalization in imitation learning involves employing image or 3D data augmentation techniques, akin to those used in computer vision. 
These techniques encompass random shifts, color adjustments, and rotations \cite{peract,laskin2020reinforcement,hansen2021generalization,yarats2021mastering}. 
Sim-to-real transfer methods utilize advanced simulation environments for domain randomization and preliminary policy training before real-world application~\cite{nguyen2018transferring,ho2021retinagan,memmel2023asid}. 
Recent developments in extensive vision and language models present a novel path to generalization: pretrained on vast image and language datasets, these models offer potentially more robust representations for robotic manipulation or can even directly inform actions~\cite{ahn2022can, huang2023voxposer, chen2023genaug, rt2}.

\section{\benchname}
\label{sec:bench}


\benchname\ is a comprehensive simulation benchmark, built by extending RLBench, consisting of 20 diverse robotic manipulation tasks, each enabled with \numvariationfactors\ \variationfactors\. We base off of RLBench as it provides a variety of realistically useful tasks, with a scripted demonstration generation framework, broad variance in their task horizon (i.e. the number of controller steps required to complete the task) and primitive actions (such as pick, place, open, close, turn, and slide). We define \variationfactors\ as scene properties, such as object color or lighting conditions. These properties can be changed to cause data distribution shifts at test-time such that the input distribution changes $p(x_{test}) \neq p(x_{train})$, but the conditional probability of action distribution remains the same 
$p(y_{test}|x_{test}) = p(y_{train}|x_{train})$, as the underlying task does not change. This form of distribution shift in Out-of-Distribution (OoD) generalization research is referred to as covariate shift~\cite{hupkes2023taxonomy}.

With \benchname, we proposed over 20,371 unique task instances from a list of 20 tasks. The tasks are also categorized into three tiers of difficulties based on the task horizon, which inherently makes the tasks harder due to compounding error in BC \cite{ross2011reduction}. The detailed breakdown of the number of unique instance per variation is also shown in Figure~\ref{fig:benchmark-data}.

We describe our task selection strategy, \variationfactors\ category and implementation in the following subsections. Thereafter, we describe the extension of \benchname\ in the real-world for 4 tasks replicated from the simulation. We finally propose the \benchname\ Challenge and explain the training and evaluation procedure expected for leaderboard participation compliance.

\subsection{Methodology for Task Selection}

We curate the task list for \benchname\ from the default suite of 100 tasks in RLBench.
This selection ensures the feasibility of generating waypoints to facilitate task execution after incorporating our \variationfactors. Our methodology involved discerning overlays within certain tasks, leading to the inclusion of tasks that require a versatile spectrum of primitive actions. This spectrum encompasses tasks ranging from straightforward ones, requiring fewer than 100 steps (e.g., \texttt{open drawer}), to more intricate challenges like \texttt{empty dishwasher}, which may exceed 1000 steps. Figure~\ref{fig:benchmark-data} shows the complete list of tasks classified into zones of complexity based on the horizon of the tasks.

\subsection{Perturbation Factors} 
We create \numvariationfactors\ \variationfactors\ and apply each of them to the above 20 tasks where compatible. We categorize them as follows:
\paragraph{Manipulation object (\texttt{MO}) perturbation} \texttt{MO} is a task-relevant object that is directly manipulated or interacted with by the robot. For instance, in \texttt{put wine in rack} task, the `wine bottle' is the manipulation object. \texttt{MO} variations include \texttt{MO\_Color}, \texttt{MO\_Texture}, \texttt{MO\_Size}.

\paragraph{Receiver object (\texttt{RO}) perturbation} \texttt{RO} is a task-relevant object that is not directly interacted with by the robot, for example, the `rack' in \texttt{put wine in rack} task. \texttt{RO} variations include \texttt{RO\_Color}, \texttt{RO\_Texture}, \texttt{RO\_Size}.

\paragraph{Background perturbation} Factors that do not relate to task-relevant objects, but are background characteristic of the scene. These variations include \texttt{Light\_Color}, \texttt{Table\_Color}, \texttt{Table\_Texture}, \texttt{Distractor} objects, \texttt{Background\_Texture} of the walls, and \texttt{Camera\_Pose}.

\paragraph{Physical perturbation}
Factors that affect physical properties of the objects involved in the task, such as, \texttt{Object\_Friction} where the task involves sliding of an object, and \texttt{Object\_Mass} where the gripper needs to adapt to force required for moving the object.

\benchname\ supports applying one or more \variationfactors\ defined above in the same scene and study its effect at test-time. 

\subsection{Implementation of Perturbation Factors}
\label{subsec:mass}
Following RLBench, our implementation utilizes \texttt{PyRep} \cite{james2019pyrep}, a low-level API, to interact with the underlying \texttt{CoppeliaSim} \cite{coppeliaSim}
simulator. The \texttt{PyRep} API allows the control of simulator properties such as color, texture, scaling, and pose of the objects in the scene.
We implement \numvariationfactors\ \variationfactors\ (shown in
Figure~\ref{fig:challenge}: part 3) as an extension of the RLBench task benchmark. We expose the configuration of supported perturbations via configuration files written in \texttt{YAML}. Our implementation is easily extensible for other researchers to build on and edit the benchmark, for adding new \variationfactors\ or new tasks, along with the ease of configuring any combination of \variationfactors\ and their parameters, where compatible.

To implement texture or color perturbations, we randomly sample a texture or color from our curated set. We provide a set of \numtextures\ textures and \numcolors\ colors. These assets were used for implementing \texttt{MO\_Color}, \texttt{MO\_Texture}, \texttt{RO\_Color}, \texttt{RO\_Texture}, \texttt{Table\_Color}, \texttt{Table\_Texture}, and \texttt{Background\_Texture}. To implement size perturbations (\texttt{MO\_Size} and \texttt{RO\_Size}), we sample a scaling factor from a continuous range, specified for each object that supports this factor. For instance, the range for \texttt{MO\_Size} in the task \texttt{basketball\_in\_hoop} is $[0.75, 1.25]$, differs from that in task \texttt{hockey} $[0.95, 1.05]$, as this parameter is quite dependent on the conditions of the objects in the scene. We include the remaining task parameters in the Appendix. The waypoints get re-scaled with the object, and if not, we reposition them proportionally with respect to the object center, while ensuring that RLBench's scripted demonstration generation from waypoints remains functional. The \texttt{Distractor} objects are sampled from a set of \numdistractors\ object models taken from the YCB Object Dataset~\cite{calli2015YCBdataset} and converted to \texttt{CoppeliaSim} compatible \texttt{.ttm} format. We utilize predefined object spawn boundaries to place these objects on table-top or on another object. To modify \texttt{Light Color}, we randomly sample RGB values from our specified range of $[0.0, 0.0, 0.0]$ to $[0.5, 0.5, 0.5]$, and apply it to all 3 directional lights surrounding the scene. We perturb \texttt{Camera Pose} for 3 cameras --- front, left shoulder, and right shoulder --- by changing their positions and orientations in Euler angles, sampled from ranges $[-0.1, -0.1, -0.1]$ to $[0.1, 0.1, 0.1]$ and $[-0.05, -0.05, -0.05]$ to $[0.05, 0.05, 0.05]$ respectively. 
\texttt{Object\_Friction} is implemented by changing the friction coefficient of the object with a value sampled from the range $[0.75, 1.0]$. \texttt{Object\_Mass} changes the mass of objects with a value sampled from a given range, where the ranges are task dependent (provided in Appendix).
We provide our texture, color, and object model assets with the benchmark code, which can also be augmented easily for additional assets, as required.

Some \texttt{MO} and \texttt{RO} perturbations do not apply to all the tasks, due to two main reasons. \texttt{RO} perturbations do not apply to tasks when there is no \texttt{RO} object, for instance, \texttt{open drawer} task. Additionally, \texttt{PyRep} doesn't support application of surface texture or scaling for objects made up of compound shapes, such as the `dishwasher' in \texttt{empty dishwasher} task.

\subsection{Real-World Tasks and Perturbations}

For the real-world extension of \benchname, we implement \variationfactors\ akin to those in simulation. We create real-world mirrors for 4 RLBench tasks: {\texttt{insert onto square peg}, \texttt{slide block to target}, \texttt{scoop with spatula},  and \texttt{setup chess}}. To ensure replicability of our real-world benchmark extension, we created these four tasks with 3D-printed objects, identical to those in the RLBench tasks, with the various variant factors to support the perturbations. We open-source our 3D-printed object models, which are inexpensive to print, to facilitate reproduction of our real-world tasks and their \variationfactors.

\begin{figure*}[t]
  \centering
  \includegraphics[width=\linewidth]{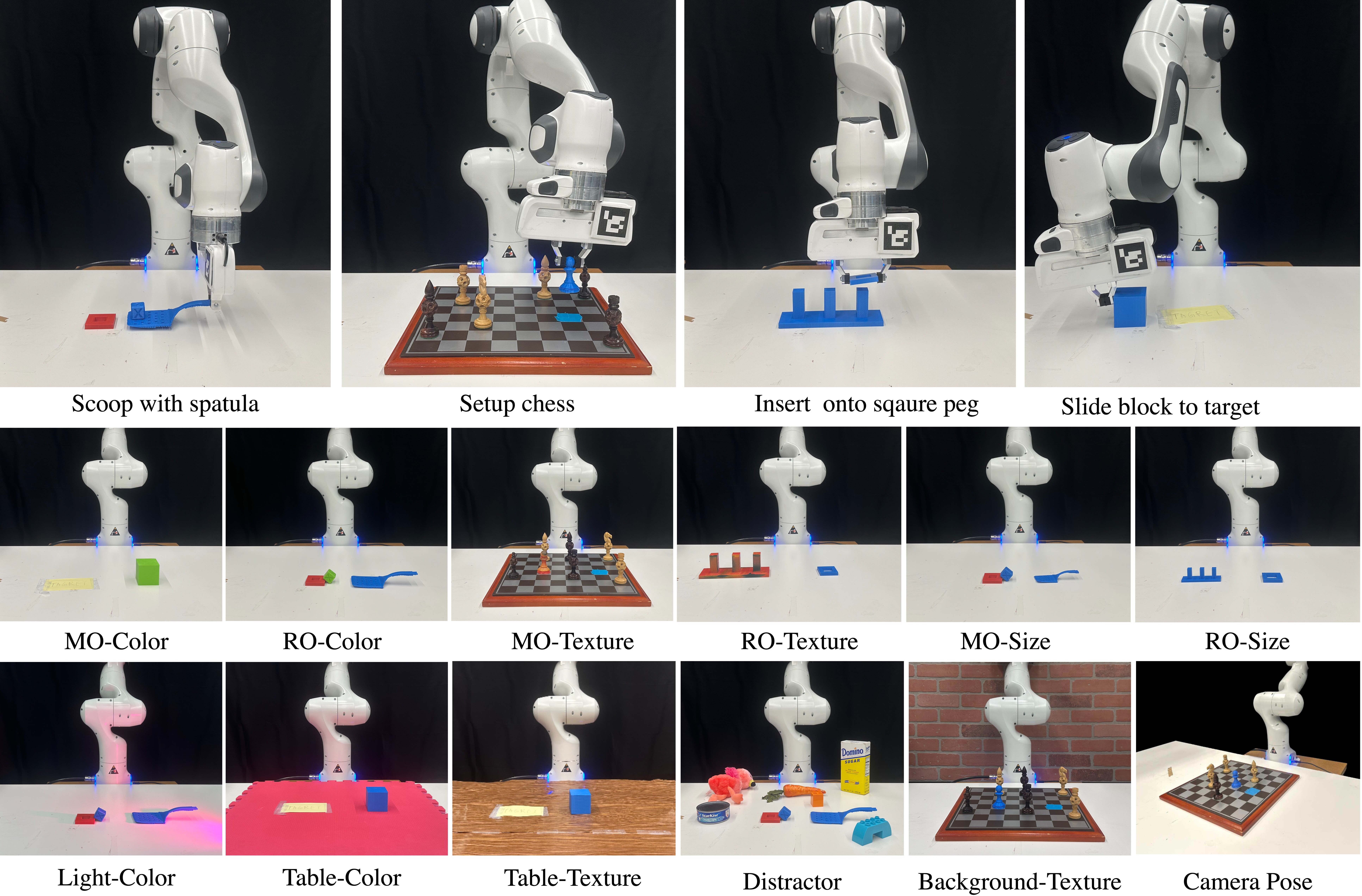}
  \caption{\textbf{Real-World training tasks and their evaluation time perturbations.} A PerAct agent, trained using real-world demonstrations for the four tasks shown, was tested on real-world \variationfactors. This evaluation involved perturbing factors similar to the procedural benchmark in the simulation. } 
  \label{fig:real-world-tasks}
\end{figure*}

Our real-world experiments utilized a Franka Panda robot arm, replicating the setup in RLBench, for both collecting training data and evaluation on the \variationfactors. To create size perturbation (\texttt{MO\_Size} and \texttt{RO\_Size}), we 3D printed identical manipulation objects in 2 additional sizes. For object color and texture perturbation (\texttt{MO\_Color}, \texttt{RO\_Color}, \texttt{MO\_Texture} and \texttt{RO\_Texture}) we 3D printed the objects in two alternate colors and textures. For \texttt{Table\_Color}, \texttt{Table\_Texture}, and \texttt{Background\_Texture}, we use two different sets of table mats or wallpapers to mirror these \variationfactors\ in the real-world. For \texttt{Camera\_Pose}, we re-calibrate the front camera at two different spots that differs from the camera pose set during training data collection. Lastly, the \texttt{Light\_Color} was simulated with a dynamically color-changing spotlight. 
Finally, to introduce \texttt{Distractor} objects, we incorporated additional random tabletop objects into the scene. We present our real-world setup in Figure~\ref{fig:real-world-tasks}.

\subsection{\benchname\ Challenge}

We propose \benchname\ Challenge to enable development of generalizable Behavior Cloning (BC) models for robotic manipulation tasks. As shown in Figure~\ref{fig:challenge}, the challenge involves four key phases: 1) Participants generate a standard training dataset for 20 tasks with 100 demonstrations each, without \variationfactors. 2) Participants train their BC models using this standardized dataset.
3) The models should evaluate over a fixed 25 episodes set of each of the \numvariationfactors\ \variationfactors. 4) Models are ranked on a leaderboard based on the percentage change in their performance across these \variationfactors. Demonstrations data for training and testing can be generated via scripted experts from RLBench. Each task is equipped to instantiate object pose-variated episodes ensuring an inexhaustible supply of task-specific demonstrations. Demonstrations are collected automatically through motion planners that navigate through manually defined waypoints.

As we show in our results, evaluation in the simulated benchmark aligns well with that on our reproducible real-world mirror. Therefore, participants can expect similar generalization in real-world when participating in \benchname\ Challenge in simulation. Participants can reproduce and evaluate on the real-world part of benchmark, however, submitting real-world results to the leaderboard will remain optional.

\section{Experiments}
\label{sec:exp}
In this section we define our problem formulation for baseline training, followed by describing the baselines methods and their respective training logistics. Thereafter, we elucidate \benchname's standard training and evaluation protocol. Finally, we describe our real-world setup and its training details.

\subsection{Dataset and Problem Formulation}
The problem is to learn action prediction from robot's observation and the language instruction. The training dataset of demonstration consists of N trajectories, $\tau_i = \{(o_j , a_j, p_j), l\}^T_{j=1}$ where $o$ is the observation and $a$ is a continuous robot arm action. An action $a_j$ is the 6-DoF gripper pose and it's open or close state, an observation $o_j$ is a set RGBD images from a given number of cameras, and robot arm's proprioception $p_j$ is the arm's current pose, at time step $j$. Each trajectory is paired with a template-generated English language instruction $l$. 

Following recent SotA ~\cite{arm, c2farm, peract, rvt}, we use keypoint-based action prediction instead of predicting continuous 7-DoF actions. The keypoint actions are discovered using intuitive heuristics, such as instances where the arm’s joint velocities are close to zero, and whether the gripper's open state has changed.

\subsection{Baselines}
We study \numsotamodels\ SotA baselines, including one zero-shot open vocabulary model (\texttt{VoxPoser}),  two 2D learning models (\texttt{R3M-MLP}, \texttt{MVP-MLP}) and two 3D learning models (\texttt{PerAct}, \texttt{RVT}). We choose these methods as baselines as they establish themselves as strong robot learning methods. They are also diverse in their approach, allowing us to study the effect of aspects such as, pretraining and 2D vs 3D based learning. For all baselines, the language is encoded using a frozen CLIP~\cite{clip} model.

\subsubsection{2D learning models} \texttt{R3M-MLP} and \texttt{MVP-MLP} use pretrained visual encoders, pretrained on out of domain task-agnostic real-world images. MVP~\cite{mvp}, a ViT-Base model with 86M parameters, learns representations with 4.5M in-the-wild images on task-agnostic real world data using a self-supervised masked reconstruction objective. R3M~\cite{r3m}, a ResNet-50 model with 23M parameters, learns representation using egocentric human videos with captions~\cite{Ego4D2022CVPR} via video-language contrastive and temporal loss objectives. Both representations have been shown to be effective for downstream task adaptation using RL or BC, via an MLP action prediction head, both in simulated and real world settings. We adapt these pretrained encoders similarly by freezing the encoders and adding an MLP prediction head with $\sim$3M trainable parameters. We train both models with batch size 32 for 300k training iterations. The input to the model is 4 camera RGB views encoded by their respective pretrained train visual encoder (no depth, as per prior work's use case), encoded language instruction, and proprioception. Following the prior work~\cite{r3m,mvp}, we predict raw 7-DoF keypoint pose of the robot arm in continuous space. 

\subsubsection{3D learning models} \texttt{PerAct} is a transformer-based robotic manipulation BC model that takes tokenized voxel grid and language instruction as the input, to predict discretized voxel grid translation point, discrete rotation in Euler angles, and gripper's binary (open/close) state. \texttt{PerAct} works with 3D voxel grid tokens, akin to visual patch tokens or language tokens in vision or language transformers. Following the original implementation, we use a voxel grid of size $100^3$, corresponding to an actual volume of 1.0m$^3$. The patch tokens of size $5^3$ are encoded via a 3D convolution layer with kernel-size and stride of 5, resulting in $20^3=8000$ voxel observation tokens. Actions are discretized via voxelized keypoint-based action prediction. The actions are then predicted as the next-best voxel that is closest to the center of the gripper fingers for the next translation pose. Rotation pose is discretized into bins of $5^o$ increments. The input to the model is the encoded language instruction, proprioception, and 4 camera RGBD views, which gets preprocessed into a voxel grid with voxel occupancy and RGB channels. The model has $\sim$33M trainable parameters. We train this model with batch size 16 for 300k training iterations.

\texttt{RVT} is a multi-view transformer-based robotic manipulation BC model that uses tokenized image patches and CLIP-encoded language instruction tokens as input to predict keypoint actions as translation heatmaps, discretized rotation in Euler angles, and gripper's binary state. \texttt{RVT} re-renders the captured RGBD views from new virtual camera views via constructing a 3D point cloud. This procedure decouples the camera images from the images fed to the model, as well as allows generating more viewpoints unrestricted by real-world constraints. The transformer attends over language instruction, re-rendered views, and robot's proprioception to predict actions. The model predicts heatmaps for each input view, which is then back-projected to a discretized set of 3D points densely populating the robot's workspace, out of which the point with the highest score is chosen as the next translation point. Rotation and gripper state prediction remain the same as \texttt{PerAct}. We train \texttt{RVT}, with $\sim$36M trainable parameters, on our 20 tasks with batch size 24 for 100k iterations, following its original configuration. 


\subsection{Zero-Shot manipulation model using Large Pretrained World Models} \texttt{VoxPoser}\cite{huang2023voxposer} is a formulation that aims to extract affordances and constraints using LLMs. Through code, it composes 3D value maps in observation space to guide robotic interactions. We utilized their RLBench implementation for VoxPoser, providing variation descriptions from each demonstration as input language text. We manually annotated all corresponding RLBench\cite{james2020rlbench} objects with their respective object names mentioned in the variation descriptions. We conducted a zero-shot evaluation of \texttt{VoxPoser} using \benchname's evaluation protocol, without any training involved.

\subsection{Training and Evaluation Protocol}
We apply \benchname\ training and evaluation to each of the above models. We train with 100 demos per task without any of \benchname\ perturbations applied. However, we do apply the default RLBench task variations in the training, i.e., changing language instruction and task target --- for instance, \texttt{open drawer}'s RLBench variations include \texttt{open bottom drawer, \texttt{open middle drawer}, \texttt{open top drawer}} --- to maintain the original baseline training settings. 

For consistent evaluation, we generate training and test data once and use the last checkpoint for each of the above trained baselines, and evaluate on each of the \variationfactors. We fix each task to the default RLBench task variations (for instance, \texttt{open bottom drawer} in the above example) in order to closely evaluate the effect of our applied perturbations. However, we do not fix the object pose variations across our test episodes. We refer to the default RLBench variation without any \variationfactors\ applied as \texttt{No Perturbation} test set. In addition, we also analyze all perturbations activated together (\texttt{All Perturbations}). That makes \benchname\ test sets 235-strong, with 25 episodes per test set. A test episode is successful if the model completes the task fully. We report the average success rate for each test set, further averaged across tasks, referred to as task-averaged success rate hereon. We include per task performance in the Appendix. We report results with one training seed and one evaluation seed over the benchmark per baseline model.

\begin{figure*}[t!]
    \centering
    \includegraphics[width=\linewidth]{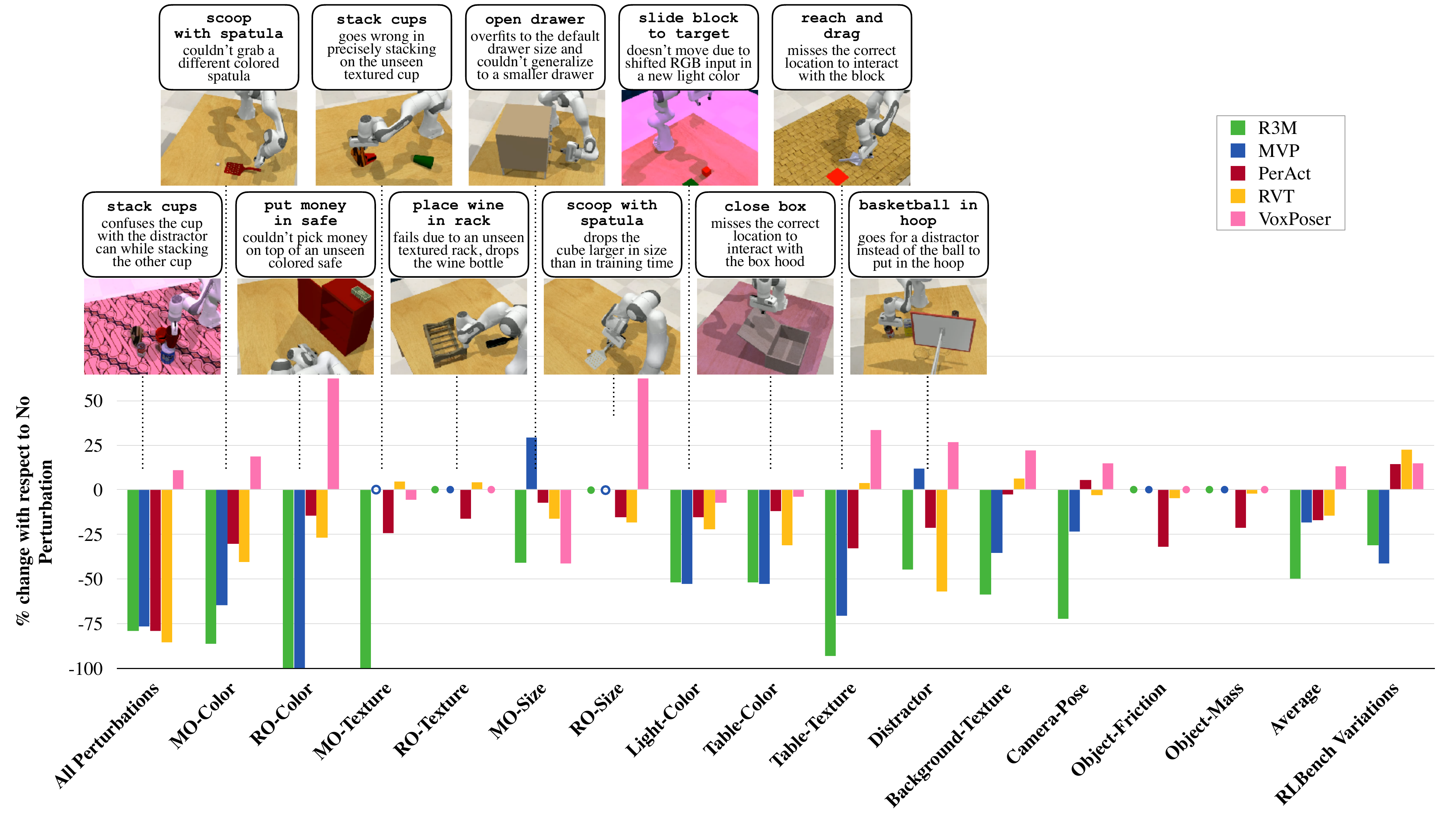}
    \caption{\textbf{Task-averaged success rate \% change for 4 baseline models on \variationfactors, compared to \texttt{No Perturbation} test set}. We report the evaluation with \texttt{All Perturbations} enabled, followed by each individual factor, average of all individual factors, and on RLBench variations (that is sampled from the same distribution as the training set). The images on top show failure examples for each factor with captions explaining the failure.
    $\bullet$ indicates undefined value when the corresponding \texttt{No Perturbation} task averages are also 0. $\circ$ indicates 0\% change with respect to \texttt{No Perturbation} task average. 
    }
    \label{fig:results1}
\end{figure*}

\subsection{Real-World Setup}
In our real-robot experiments, we employed a Franka Panda manipulator equipped with a parallel gripper for data collection and evaluation. For perception, a front-facing Kinect-2 RGB-D camera was utilized. We collected real-world demonstrations using an HTC Vive controller, gathering 5 demonstrations for each of the 4 tasks. 
To facilitate comparative analysis, we trained a \texttt{PerAct} model~\cite{peract} with $\sim$33M parameters on these real-world demonstrations for 200k iterations with a batch size of 1. In a similar vein, we trained another multi-task instance of \texttt{PerAct} in the simulated environment, focusing on the same four tasks. This simulation model, also comprising $\sim$33M parameters, was trained over 50k iterations with a batch size of 4. For consistency in evaluation between simulation and the real-world, we evaluated both models on all \variationfactors\ and \texttt{No Perturbation} test sets, each with 10 episodes, for 3 separate runs. 



\section{Results}
\label{sec:results}

We report our results as task-averaged success rate of different baselines on \benchname\ and draw insights based on which \variationfactors\ affect which kind of baseline. We also perform an upper bound training ablation when training and testing with \texttt{All Perturbations} enabled. Lastly, we report our simulation and real-world benchmark alignment analysis based on model success rates on the \variationfactors.

\subsection{Performance of different baselines on \benchname}

\begin{figure*}[t]
  \centering
  \includegraphics[width=\linewidth]{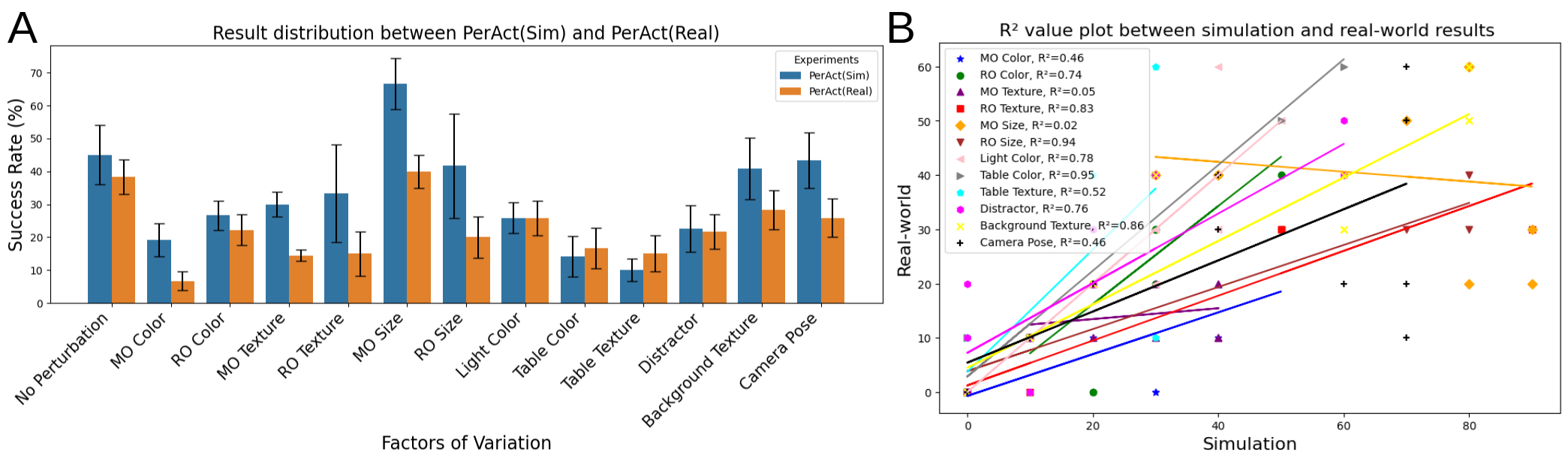}
  \caption{\textbf{Real robot results and alignment analysis on \variationfactors\ across 4 tasks.} A) The plot illustrates the empirical performance of two models, one trained using real-world data and the other with simulated data, each tested across 10 episodes and 3 runs in their respective environments. Additionally, it displays a uniform distribution of standard deviation between the models, highlighting which \variationfactors\ align more strongly with empirical success rates. B) To examine the correlations between simulation and real-world results for each perturbation independently, we have plotted a scatter chart. This chart includes data points from one run for each task. We calculated the $R^2$ value for each task and illustrated their respective best-fit lines on the chart. 
  } 
  \label{fig:real-world-results}
\end{figure*}

We report absolute task-averaged success rates in Figure~\ref{fig:results} for all baselines and \variationfactors. 
All polar axes are comparable, except \texttt{MO\_Texture}, \texttt{MO\_size}, \texttt{RO\_Color}, \texttt{RO\_Texture}, and \texttt{RO\_Size} axes that have different task averages, since they don't apply to all 20 tasks (more details reported in the Appendix).
We also report the above results as percentage change with respect to \texttt{No Perturbation}, averaged across 20 tasks, along with qualitative failure cases associated with each of the \variationfactors\ in Figure~\ref{fig:results1}. For factors that were not applicable or infeasible in simulation on some tasks, we compare their averages only with corresponding task's \texttt{No Perturbation} case on which that perturbation was applied and evaluated. Applying \texttt{All Perturbations} in the same scene influences all the models significantly, leading to $\geq$75\% decrease in performance. What \variationfactors\ are the most affecting?

\textbf{For 2D-models (\texttt{R3M-MLP} and \texttt{MVP-MLP}), we observe that object and light color, texture, and camera pose are the most affecting factors.} Since these models are trained end-to-end with RGB inputs, and the color or texture related perturbations shift the input space, thereby affecting the output space as well. Moreover, training with specific \texttt{Camera\_Pose}s when using RGB as input also affects the performance when camera poses are perturbed. We observe that \texttt{MVP-MLP} does better or is not affected in presence of \texttt{Distractors}, which may be due to the real world pretraining on cluttered scenes. This result indicates value in pretraining on real-world data.

\textbf{For zero-shot manipulation models using Large Pretrained World Models, we observed that the system demonstrates robust generalization capabilities across various conditions, particularly excelling in tasks where it is predisposed to succeed.} Specifically, for the two tasks in which \texttt{VoxPoser} excels, it maintains consistent performance across all variants. For example, in the task \texttt{slide\_block\_to\_target}, the performance difference between the \texttt{No Perturbation} scenario and the average of all perturbations is a mere 3.21\% relative to the \texttt{No Perturbation} performance. This aligns with our expectation that leveraging large pretrained world models enables the recognition of significant changes in environments and object perturbations.

\textbf{For 3D-models (\texttt{RVT} and \texttt{PerAct}), we observe that the most affecting factors are color-related including object, table and light colors as well as presence of \texttt{Distractors}}, while other factors cause a smaller performance decrease. Since \texttt{RVT} and \texttt{PerAct} are both trained end-to-end with RGB images or voxel grid with RGB channels, the color perturbations remain challenging for these models as well. These models lack any real-world pretraining, thus, the presence of \texttt{Distractors} puts the scene out of distribution, significantly affecting their performance. \textbf{We observe that these model are robust to changes in \texttt{Camera\_Pose}, because they do not directly learn on captured view.} They instead preprocess the input RGBD views into a voxel grid or re-rendered novel views. For these models, while each factor doesn't lead to a very significant effect on their performance, all factors combined in one scene (\texttt{All Perturbations}) cause a significant decrease. 
On physical perturbations \texttt{RVT} performs better than \texttt{PerAct}, perhaps because modelling in \texttt{RVT} is more robust for keypoint prediction than \texttt{PerAct} under these perturbations. Physical perturbation results for other models are inconlcusive as they cannot perform the tasks that support these perturbations.

\textbf{We observe that 3D baselines are better performing generally (Figure~\ref{fig:results}), and on average much more robust to environment perturbations as compared to 2D baselines (Figure~\ref{fig:results1}).} We also observe that \texttt{RVT}, trained only with RGB views, generally gets more affected with \variationfactors\ as compared to \texttt{PerAct}, trained with complete 3D scene, notably in the case of \texttt{Distractors}. This result indicates value in learning with 3D scenes as input, for the resultant model is more robust to such environmental perturbations, as it might be learning 3D features of the objects instead of just their 2.5-dimensional projections.

\begin{figure}[th]
    \centering
    \includegraphics[width=\linewidth]{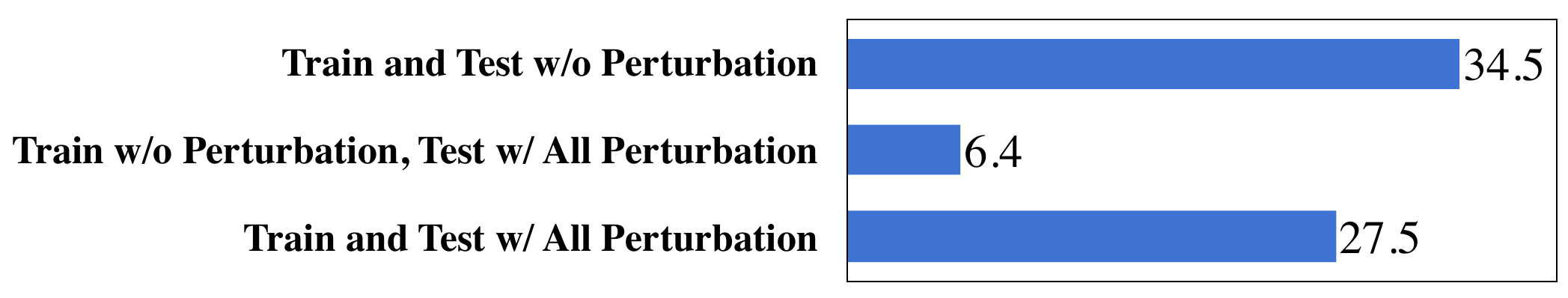}
    \caption{\textbf{Task-averaged success rate ablation for \texttt{All Perturbations}} when training and testing with or without all perturbations enabled.}
    \label{fig:upper-bound}
\end{figure}

\subsection{Training on \texttt{All Perturbations} ablation} 
\label{sub:all_perturbation}

We report results on training \texttt{PerAct} with 100 demos (with batch size 16 for 300k iterations) in \texttt{All Perturbations} setting in Figure~\ref{fig:upper-bound}. Zero-shot evaluation of a \texttt{PerAct} model trained on RLBench variations data with no \variationfactors\ enabled achieves task-averaged success rate of 6.4\% (28.1\% lower than \texttt{No Perturbations} task-averaged success rate). 
When we train the model with with \texttt{All Perturbations} enabled, the task-averaged success rate increases by 21.1\% only. 
However, the model should be able to perform the same tasks under any environmental setting for being practically deployable. This result indicates that \benchname's \variationfactors\ not only study systematic perturbations added to the environment, but also increase the difficulty of the tasks itself, even with ground truth perturbed scenes available for training.
From this ablation with \texttt{All Perturbations}
as proxy for an extreme case of factor compounding, which is
results in 28.1\% lower in success rates with respect to \texttt{No Perturbations}, hence suggesting that compounding perturbation factors has some degree of compounding effect toward models' performance.

\subsection{Real-world alignment analysis for \benchname} We first observed only a marginal difference in success rate of 6.67\% during evaluation on the  \texttt{No Perturbations} tasks between \texttt{PerAct} trained in simulation and that in real-world settings. This served as a crucial sanity check for tasks performance between the two models before advancing forward to evaluating them across the \numvariationfactors\ \variationfactors. We observed that for factors such as \texttt{MO\_Texture}, \texttt{Light\_Color}, \texttt{Table\_Color}, \texttt{Table\_Texture}, and \texttt{Distractor}, the discrepancies in performance between both models on each individual factor were marginal, remaining under 5\%.

\begin{figure*}[t]
  \centering
  \includegraphics[width=\linewidth]{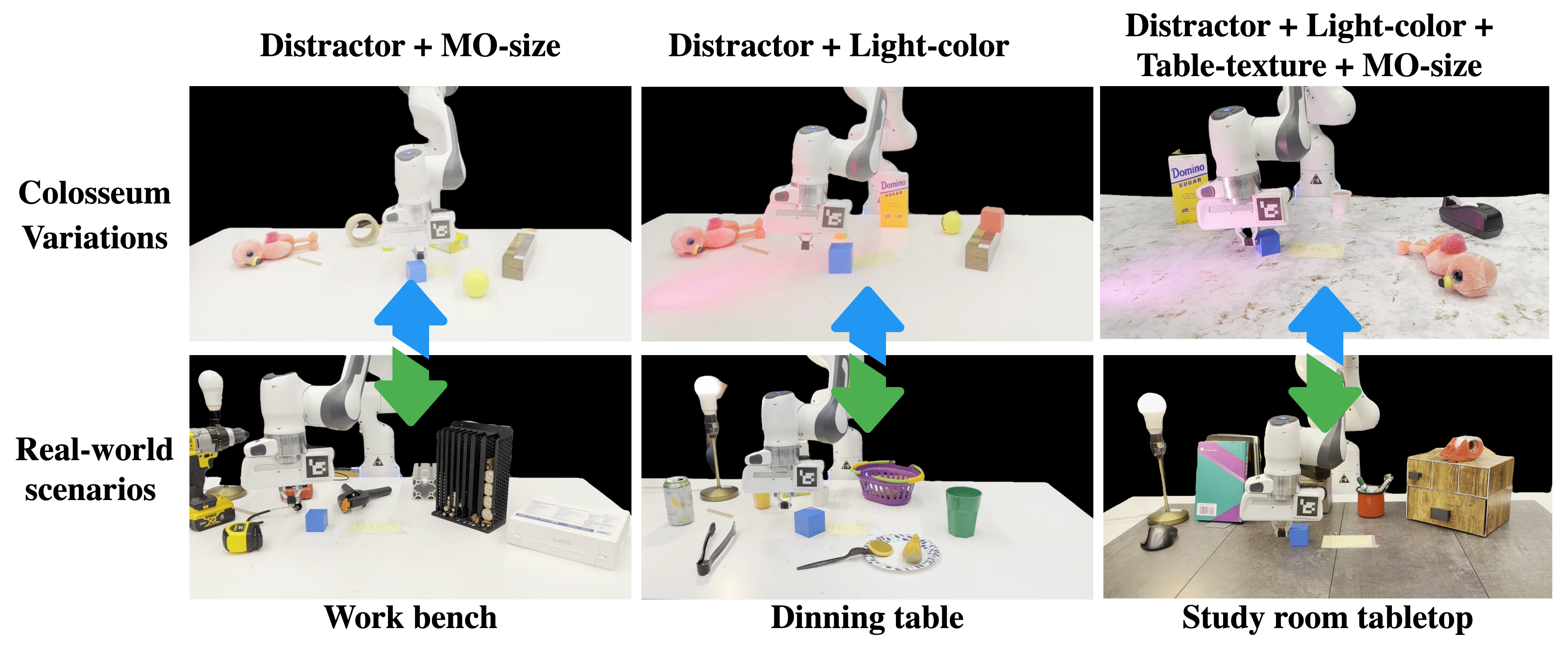}
  \caption{\textbf{Example rollouts using \benchname\ perturbations in tandem with comparable real-world scenarios.} We combine various \benchname\ variations to form three combinations of compounding perturbation scenarios, each paired with a curated real-world scene (\emph{workbench, dining table, study room table}) featuring the same set of perturbations that are naturally present in the scene, similar to the approach used by \cite{chen2023genaug}. We evaluate these combinations using \texttt{PerAct} model trained on \texttt{No Perturbation}, and aim to establish a correlation between our simplified compounding perturbation scenarios and realistic real-world scenes.
  }
  \label{fig:rebuttal}
\end{figure*}


The observed variances in other factors may stem from differences in waypoint annotations, physical robot interactions, and training data's visual distinctions.
To further investigate the correlation between simulation and reality, we used the success rate performance of each individual run for each task as a data point to calculate the coefficient of determination, or R-square values. Our results indicated that for factors like \texttt{MO\_Color}, \texttt{Table\_Texture}, and \texttt{Camera\_Pose}, there was a moderate level of correlation (0.46$\leq$R²$\leq$0.52). Conversely, factors 
such as \texttt{Background\_Texture}, \texttt{Distractor}, \texttt{Table\_Color}, \texttt{Light\_Color}, \texttt{RO\_Color}, \texttt{RO\_Texture}, and \texttt{RO\_Size} has a R² value between 0.74 and 0.94 with \texttt{Table\_Color} being the most significant as illustrated in Fig \ref{fig:real-world-results}B. These results suggest that for at least 7 out of \numvariationfactors\ \variationfactors\  there is a strong correlation between the performances of the two models, \textbf{thereby indicating a clear alignment between evaluation done on \benchname\ in simulation and in the real-world}.

Based on the results presented in Figure~\ref{fig:real-world-results}A, we also observed that for real-world experiments, \texttt{MO\_Color} exhibited a substantial decline in task success, with an 82.6\% drop, whereas \texttt{MO\_Size} demonstrated no performance reduction, instead enhancing performance by 4.34\%. Further examination of individual episodes in the real-world experiments, we observed that perturbations in \texttt{Light\_Color} could significantly alter an object's visual appearance by casting differently colored light, consequently impacting the BC model's success rate. Additionally, the \texttt{MO\_Color} perturbation frequently impeded the robot's ability to accurately predict the 6D pose for grasping the manipulation object. This finding is consistent with results from simulation and underscores a critical aspect: BC models like \texttt{PerAct}, which construct their 3D encoders from the ground up without leveraging pretrained 3D features, struggle to generalize across a wide range of object visuals. This limitation highlights the challenge in developing robust BC models capable of adapting to diverse visual environments.

\subsection{\benchname\ perturbations grounded in the real-world}
\label{subsec:real-world-ablation}
To validate that the perturbations in \benchname\ accurately mimic naturally occurring environmental or object variations in real-world scenarios, we carried out an ablation study. This study compared three realistic scenes—workbench, dining table, and study room tabletop—with corresponding perturbation combinations derived from \benchname\ as shown in Figure \ref{fig:rebuttal}. Utilizing a multitask instance of \texttt{PerAct} trained with \texttt{No Perturbation}. We evaluated all the pairs of scenarios for the task of \texttt{slide\_block\_to\_target}. Over five trials with ten episodes each, we observed a significant correlation. Notably, the combination of [\texttt{Distractor} + \texttt{MO\_Size}] resulted in an ${R}^2 = 0.75$, while the combination of [\texttt{Light\_Color} + \texttt{Table\_Texture} + \texttt{Distractor} + \texttt{MO\_Size}] achieved an ${R}^2 = 0.83$. For further details on the results and methodology, please refer to Supplementary Section IV.E. 
\section{Limitations and Future Work} 
\label{sec:future}

Currently, our leaderboard baselines only include 4 methods, which are all BC methods. In future, we plan to include RL-based methods~\cite{NeRF-RL}. In addition, we also plan to include several other baseline methods, such as, those based on diffusion~\cite{chi2023diffusionpolicy}, 3D feature feature fields~\cite{gervet2023act3d, Ze2023GNFactor}, large-scale robotics pretraining~\cite{octo_2023}, action tokenization~\cite{rt2}, and action chucking~\cite{act-aloha}. Expanding \benchname\ with these methods will unify comparing effectiveness of robot learning methods in a single leaderboard, while also providing a good starter framework for researcher to develop new methods along or beyond included baselines.

In our real-world experiments, a key limitation lies in precisely replicating the pose, orientation, and execution of tasks both in the collected training data and during evaluation. Additionally, due to resource constraints, each perturbed factor in the real-world setup was limited to only two alternate variations. As a result, the real-world findings primarily represent a comparative performance distribution between the simulation and real-world scenarios. Looking ahead, we aim to expand the number of real-world tasks, ensuring they closely mirror their counterparts in simulation. This expansion is intended to enhance the reproducibility of simulated tasks, thereby broadening benchmarking scope. 

\section{Conclusion } 
\label{sec:conclusion}
We introduced \benchname, a comprehensive benchmark designed to assess the generalization capabilities of Behavior Cloning (BC) models in robotic manipulation. 
\benchname\ systematically perturbs the task environments of the robot's workspace along an exhaustive list of axes --- including object appearance and size, lighting, physical properties of objects, background, table-top appearance, and camera pose --- both in simulation and real-world.
Through empirical studies conducted with SotA BC methods on \benchname, we identified which perturbation factors most significantly impact model's success rates on tasks they are trained to execute. 
Additionally, we demonstrated a close alignment between \benchname\ in simulated and real-world. 
To enhance reproducibility and facilitate future model evaluations in both simulated and real-world, we will open-source our resources along with the 3D printed assets. 
\benchname\ offers a platform for future research to develop and quantitatively evaluate robotic manipulation models before scaling via a unified leaderboard.

\section{Acknowledgement} 
\label{sec:ack}

Jiafei Duan is supported by the Agency for Science, Technology and Research (A*STAR) National Science Fellowship. We would also like to thank Yi Li and the members of the UW RSE Lab for their feedback on the paper.



\bibliographystyle{plainnat}
\bibliography{reference}

\clearpage
\section*{\textbf{Author Contributions}}
\label{sec:contributions}


The first three authors contributed equally to the project:
\begin{itemize}
    \item \textbf{Wilbert}: Developed the benchmark, built all the simulation perturbations, documented the benchmark, created the website, and will continue to maintain and augment the benchmark.
    \item \textbf{Ishika}: Ideated and initiated the project, worked closely with Wilbert to define the perturbations and structure of the benchmark codebase, setup and ran benchmark baseline trainings, evaluations and analysis, drafted and structured the paper, and will continue to maintain and augment the benchmark.
    \item \textbf{Jiafei}: Contributed significantly to the project idea and benchmark perturbation definitions, setup and ran real world experiments and analysis, drafted and structured the paper, created the website, and will continue to maintain and augment the benchmark.
\end{itemize}

We want to keep improving this benchmark and the framework by adding more tasks, perturbation factors, and SotA models as baselines as they come. Please reach out to add any of the above to \benchname\, and we will be happy to include your contribution and acknowledge you in \benchname\'s README updates.
\section*{\textbf{\benchname\ Appendix}} \label{sec:supp_benchmark}

\begin{figure*}[h]
  \centering
  \includegraphics[width=\textwidth]{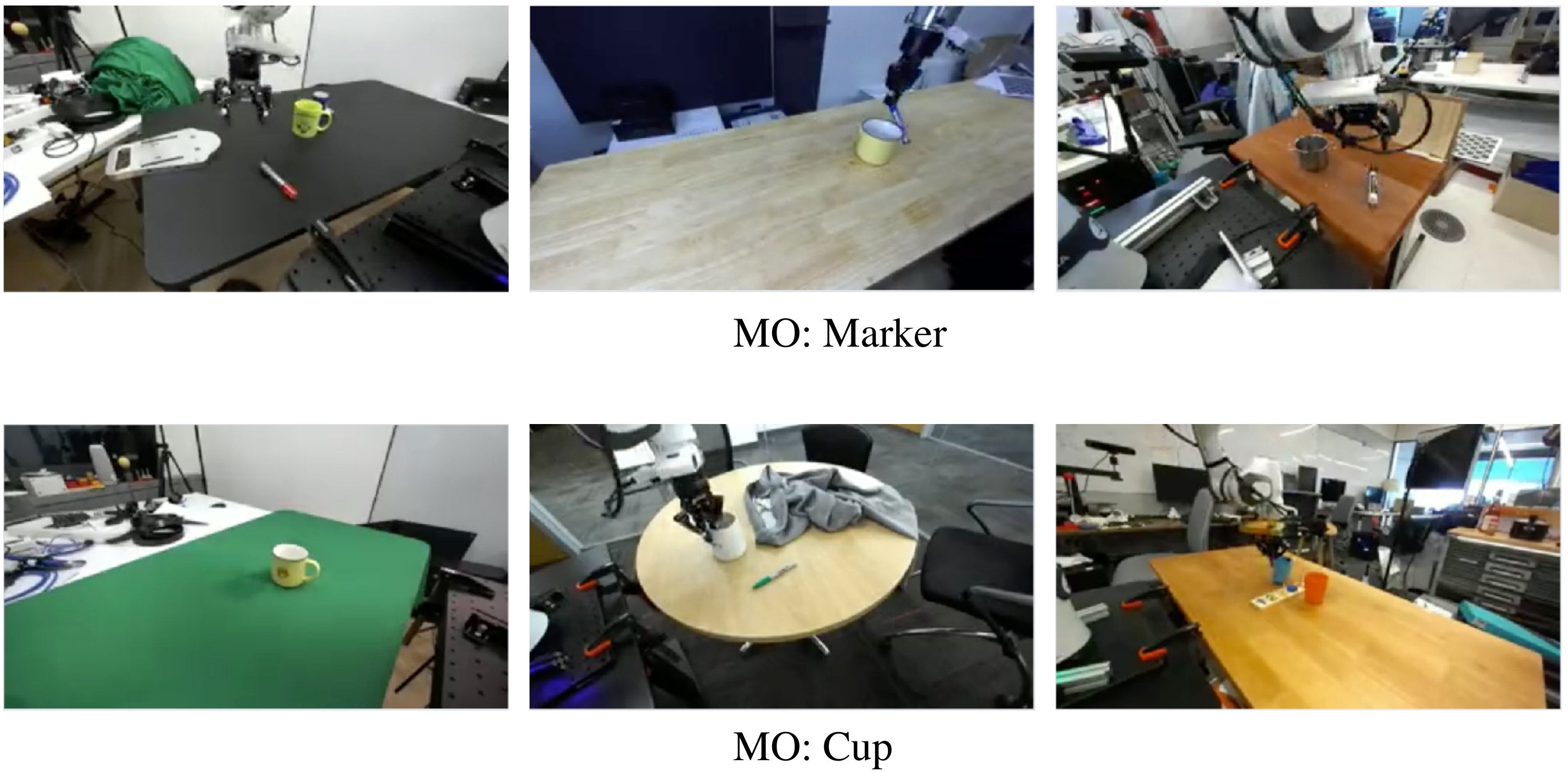}
  \caption{Dataset samples from DROID showing scene variations including \texttt{MO\_Color/Texture/Size, Light\_Color, Table\_Color/Texture, Distractors, Camera\_Pose, Background}, supporting our choice  \variationfactors.}
    
  \label{fig:survey}
\end{figure*}

\section{Perturbation Factors Selection Rationale}
We study recent real-world diverse robot datasets, such as, Open-X~\cite{open_x_embodiment_rt_x_2023}, DROID~\cite{khazatsky2024droid}, Ego4D~\cite{Ego4D2022CVPR}, and conclude that our identified factors indeed exist in these datasets. In Figure~\ref{fig:survey}, randomly sampled from DROID dataset, we can observe that \texttt{MO\_Color/Texture/Size, Light\_Color, Table\_Color/Texture, Distractors, Camera\_Pose, Background} changing across scenes. While it is not explicity reported, we can also infer that mass of cups would also change. While these factors do not cover the exhaustive list of factors that vary in the real-world, our empirical analysis shows that \benchname\ factors do affect the SoTA robot manipulation models, and hence are important to study. It is challenging to breakdown real-world into an exhaustive systematic enumeration of factors. \benchname\ is one of the first attempt towards increasing real-world task robustness for robotic manipulation via such a systematic purturbation benchmark.

\section{Simulation task Details} \label{app:task_details}
We describe each of the 20 tasks in detail, along with their RLBench variations and success condition. 

\subsection{\texttt{open drawer}}
\textbf{Filename:} $\texttt{open\_drawer.py}$ 

\textbf{Task:} Open one of the three drawers: \texttt{bottom}, \texttt{middle}, or \texttt{top}.
\textbf{Success Metric}: The prismatic joint of the specified drawer is fully extended. 

\subsection{\texttt{slide block to target}}
\textbf{Filename:} $\texttt{slide\_block\_to\_target.py}$ 

\textbf{Task:} Slide the block to square target.
\textbf{Success Metric}: Some part of the block is inside the specified target area.

\subsection{\texttt{basketball in hoop}}
\textbf{Filename:} $\texttt{basketball\_in\_hoop.py}$ 

\textbf{Task:} Pick up the basketball and put it into the hoop.
\textbf{Success Metric}: 1 basketball falls into the hoop. 

\subsection{\texttt{meat on grill}}
\textbf{Filename:} $\texttt{meat\_on\_grill.py}$ 

\textbf{Task:} Take either the \texttt{chicken} or \texttt{steak} off the rack and put it on the grill. 
\textbf{Success Metric}: The specified meat is on the grill. 

\subsection{\texttt{close box}}
\textbf{Filename:} $\texttt{close\_box.py}$ 

\textbf{Task:} Close the box.  
\textbf{Success Metric}: The revolute joint of the specified handle is at least $60^\circ$ off from the starting position. 

\subsection{\texttt{close laptop lid}}
\textbf{Filename:} $\texttt{Close\_Laptop\_Lid.py}$ 

\textbf{Task:} Close the laptop lid.
\textbf{Success Metric}: The revolute joint of the specified handle is at least $60^\circ$ off from the starting position.

\subsection{\texttt{empty dishwasher}}
\textbf{Filename:} $\texttt{empty\_dishwaser.py}$ 

\textbf{Task:} Open the dishwasher and take out the plate.
\textbf{Success Metric}: The plate has been taken out of the dishwasher.

\subsection{\texttt{reach and drag}}
\textbf{Filename:} $\texttt{reach\_and\_drag.py}$ 

\textbf{Task:} Grab the stick and use it to drag the cube on to the target square.
\textbf{Success Metric}: Some part of the block is inside the specified target area.

\subsection{\texttt{get ice from fridge}}
\textbf{Filename:} $\texttt{get\_ice\_from\_fridge.py}$ 

\textbf{Task:} Pick up the cup and push it against the ice dispenser. 
\textbf{Success Metric}: Cup pushed against the ice dispenser. 

\subsection{\texttt{hockey}}
\textbf{Filename:} $\texttt{hockey.py}$ 

\textbf{Task:} Pick up the hockey stick, and hit the ball into the goal pose. 
\textbf{Success Metric}: The ball enters into the goal pose. 

\subsection{\texttt{put money in safe}}
\textbf{Filename:} $\texttt{put\_money\_in\_safe.py}$ 

\textbf{Task:} Pick up the stack of money and put it inside the safe on the specified shelf. The shelf has three placement locations: \texttt{top}, \texttt{middle}, \texttt{bottom}. 
\textbf{Success Metric}: The stack of money is on the specified shelf inside the safe. 

\subsection{\texttt{place wine at rack location}}
\textbf{Filename:} $\texttt{place\_wine\_at\_rack\_location.py}$ 

\textbf{Task:} Grab the wine bottle and put it on the wooden rack at one of the three specified locations: \texttt{left}, \texttt{middle}, \texttt{right}. The locations are defined with respect to the orientation of the wooden rack. 
\textbf{Success Metric}: The wine bottle is at the specified placement location on the wooden rack. 

\subsection{\texttt{move hanger}}
\textbf{Filename:} $\texttt{move\_hanger.py}$ 

\textbf{Task:} Pick up the hanger and move it from one side to another.  
\textbf{Success Metric}: The hanger is successfully hooked onto the other hanger holder.

\subsection{\texttt{wipe desk}}
\textbf{Filename:} $\texttt{wipe\_desk.py}$ 

\textbf{Task:} Pick up the sponge and wipe the dust particles off the desk.
\textbf{Success Metric}: The table is being cleaned up.

\subsection{\texttt{straighten rope}}
\textbf{Filename:} $\texttt{straighten\_rope.py}$ 

\textbf{Task:} Pick up one end of the rope and move it to the nearest tape patch, and the same for the other end. 
\textbf{Success Metric}: The two patches have one side of the rope on each.

\subsection{\texttt{insert onto square peg}}
\textbf{Filename:} $\texttt{insert\_onto\_square\_peg.py}$ 

\textbf{Task:} Pick up the square and put it on the specified color spoke. The spoke colors are sampled from the full set of 20 color instances. 
\textbf{Success Metric}: The square is on the specified spoke.

\subsection{\texttt{stack cups}}
\textbf{Filename:} $\texttt{stack\_cups.py}$ 

\textbf{Task:} Stack all cups on top of the specified color cup. The cup colors are sampled from the full set of 20 color instances. The scene always contains three cups.   
\textbf{Success Metric}: All other cups are inside the specified cup.

\subsection{\texttt{turn oven on}}
\textbf{Filename:} $\texttt{turn\_oven\_on.py}$ 

\textbf{Task:} Grasp onto the knob and turn it on. 
\textbf{Success Metric}: The knob is turned on. 

\subsection{\texttt{setup chess}}
\textbf{Filename:} $\texttt{setup\_chess.py}$ 

\textbf{Task:} Pick up the odd chess pieces and put it into the start position. 
\textbf{Success Metric}: The odd one out chess piece has been placed on the designated spot. 

\subsection{\texttt{scoop with spatula}}
\textbf{Filename:} $\texttt{Scoop\_with\_Spatula.py}$ 

\textbf{Task:} Pick up the spatula and scoop up the cube. 
\textbf{Success Metric}: The cube has been successfully picked up using the spatula.

\section{Simulation details}
\begin{figure*}
    \centering
    \includegraphics[width=\linewidth]{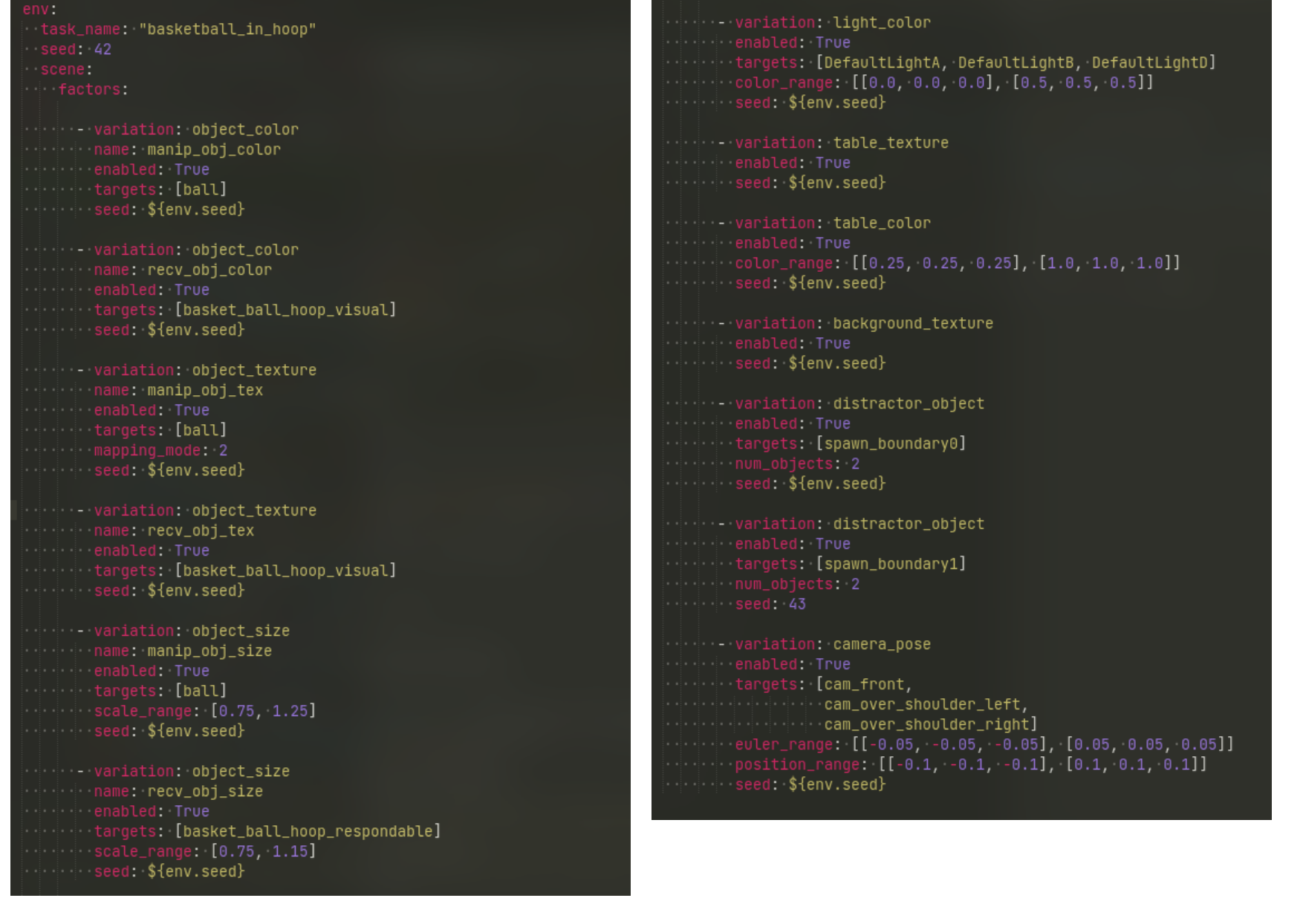}
    \caption{Sample of a \texttt{yaml} configuration file for \benchname\ for one task. This configuration file controls the application of each \variationfactors\ for this task. One or more factors can be applied at the same time in one task instance, as compatible.}
    \label{fig:app_config_file}
\end{figure*}

\begin{figure*}
    \centering
    \includegraphics[width=\linewidth]{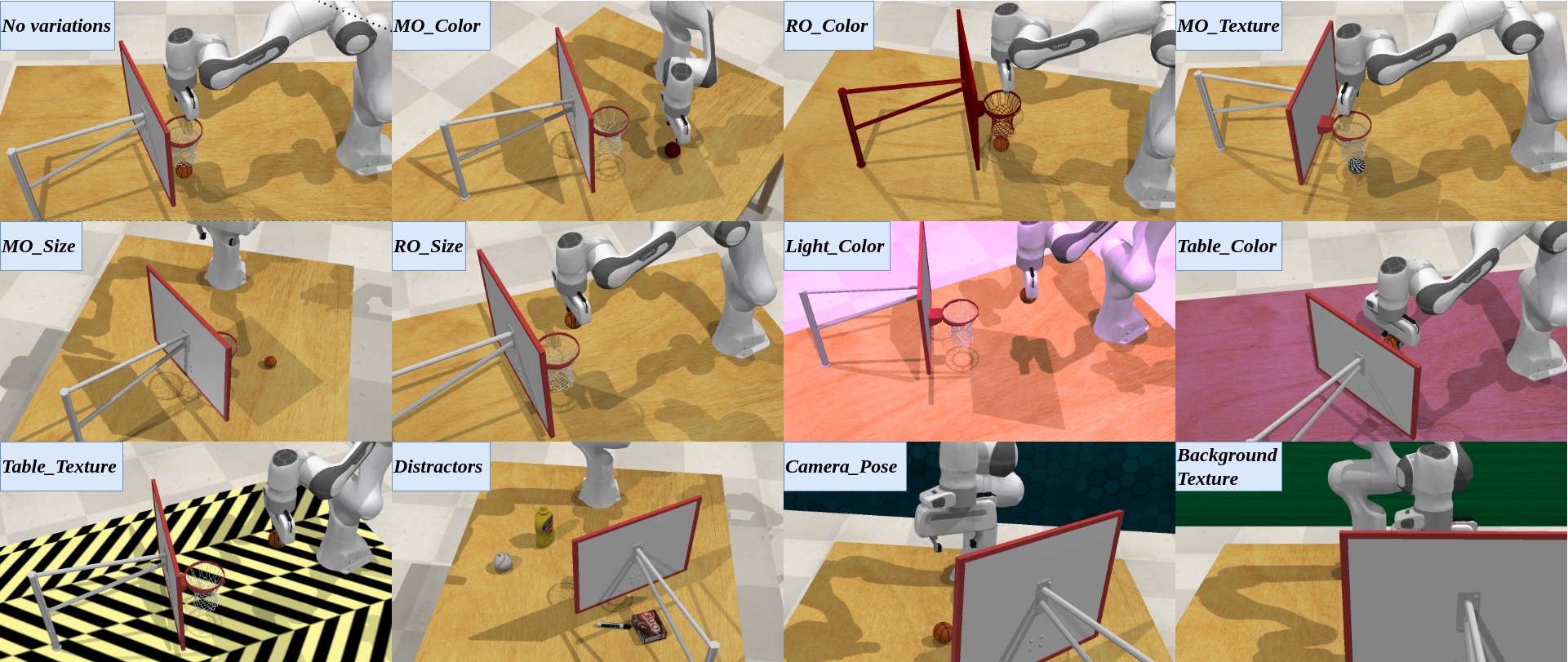}
    \caption{Perturbations for the \texttt{basketball\_in\_hoop} task}
    \label{fig:app_perturbations_basketball}
\end{figure*}

\begin{figure*}
    \centering
    \includegraphics[width=\linewidth]{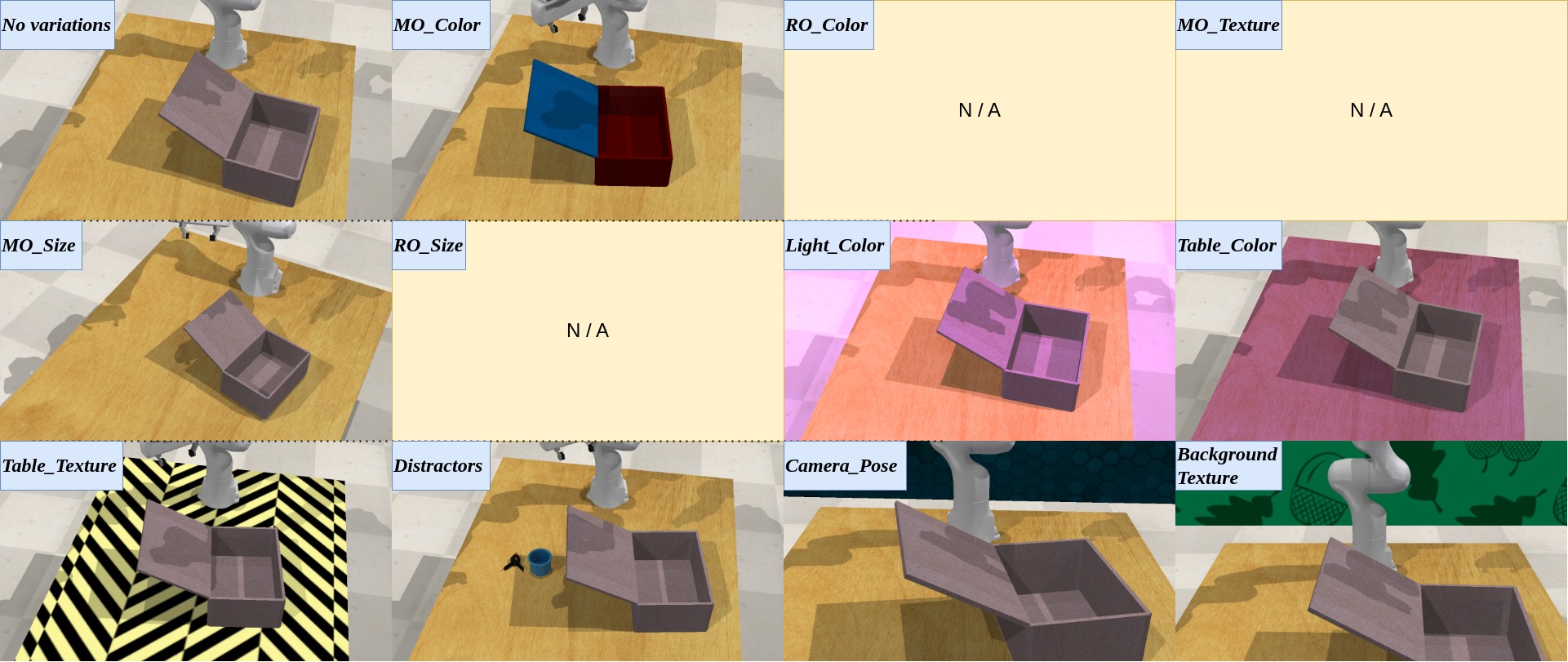}
    \caption{Perturbations for the \texttt{close\_box} task}
    \label{fig:app_perturbations_close_box}
\end{figure*}

\begin{figure*}
    \centering
    \includegraphics[width=\linewidth]{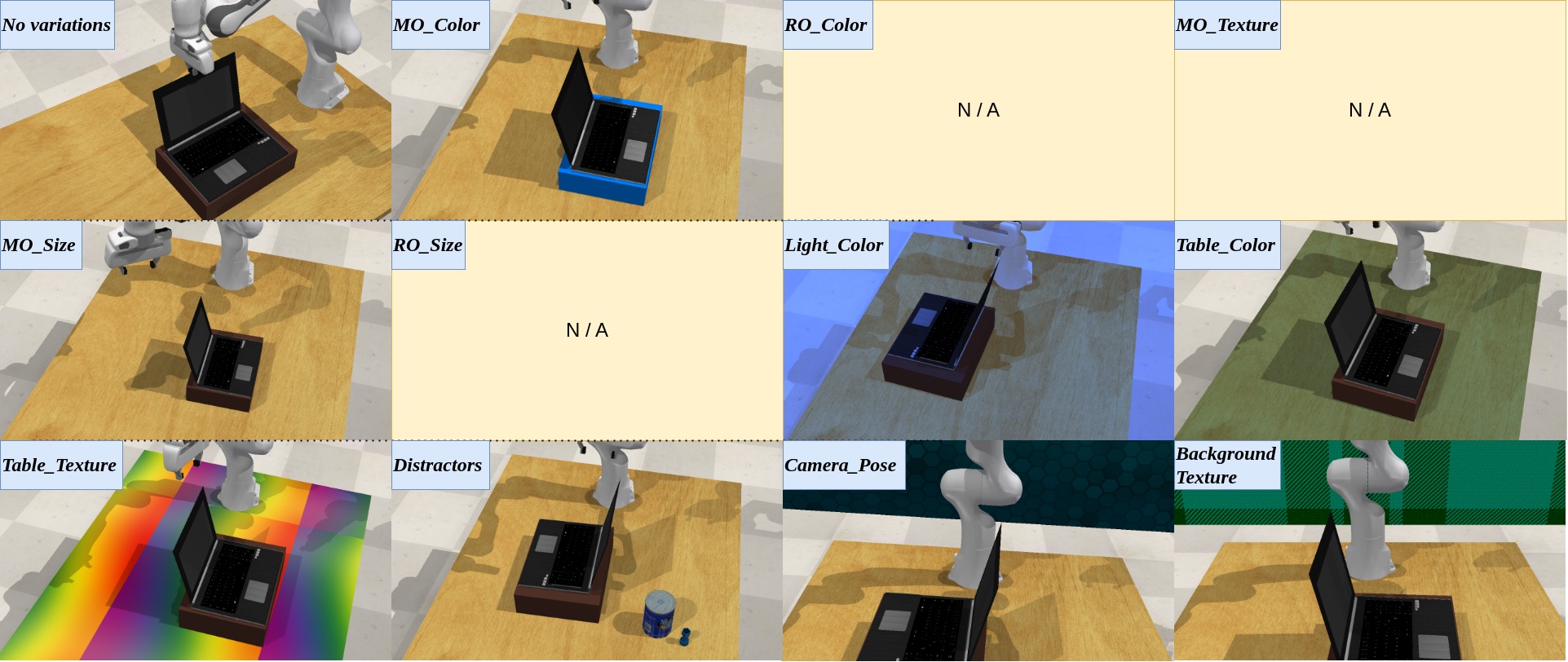}
    \caption{Perturbations for the \texttt{close\_laptop\_lid} task}
    \label{fig:app_perturbations_close_laptop_lid}
\end{figure*}

\begin{figure*}
    \centering
    \includegraphics[width=\linewidth]{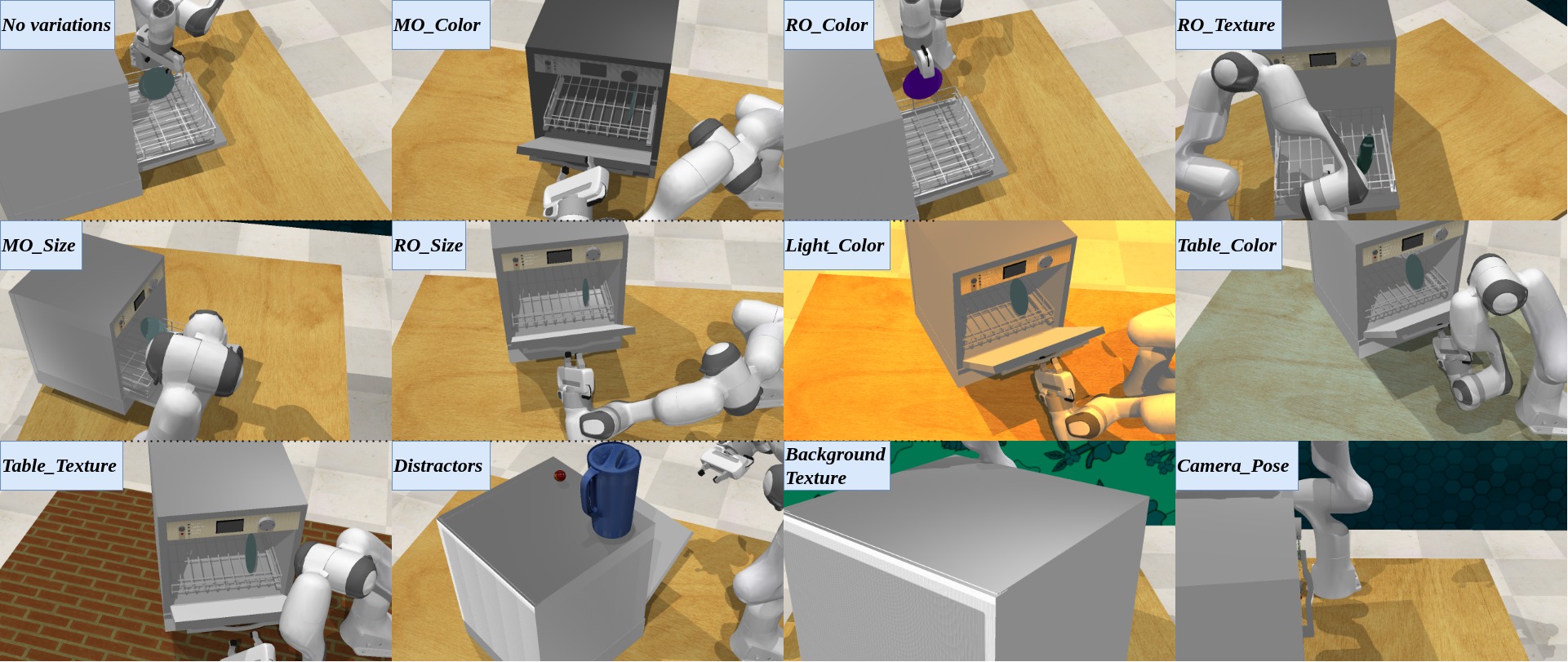}
    \caption{Perturbations for the \texttt{empty\_dishwasher} task}
    \label{fig:app_perturbations_empty_dishwasher}
\end{figure*}

\begin{figure*}
    \centering
    \includegraphics[width=\linewidth]{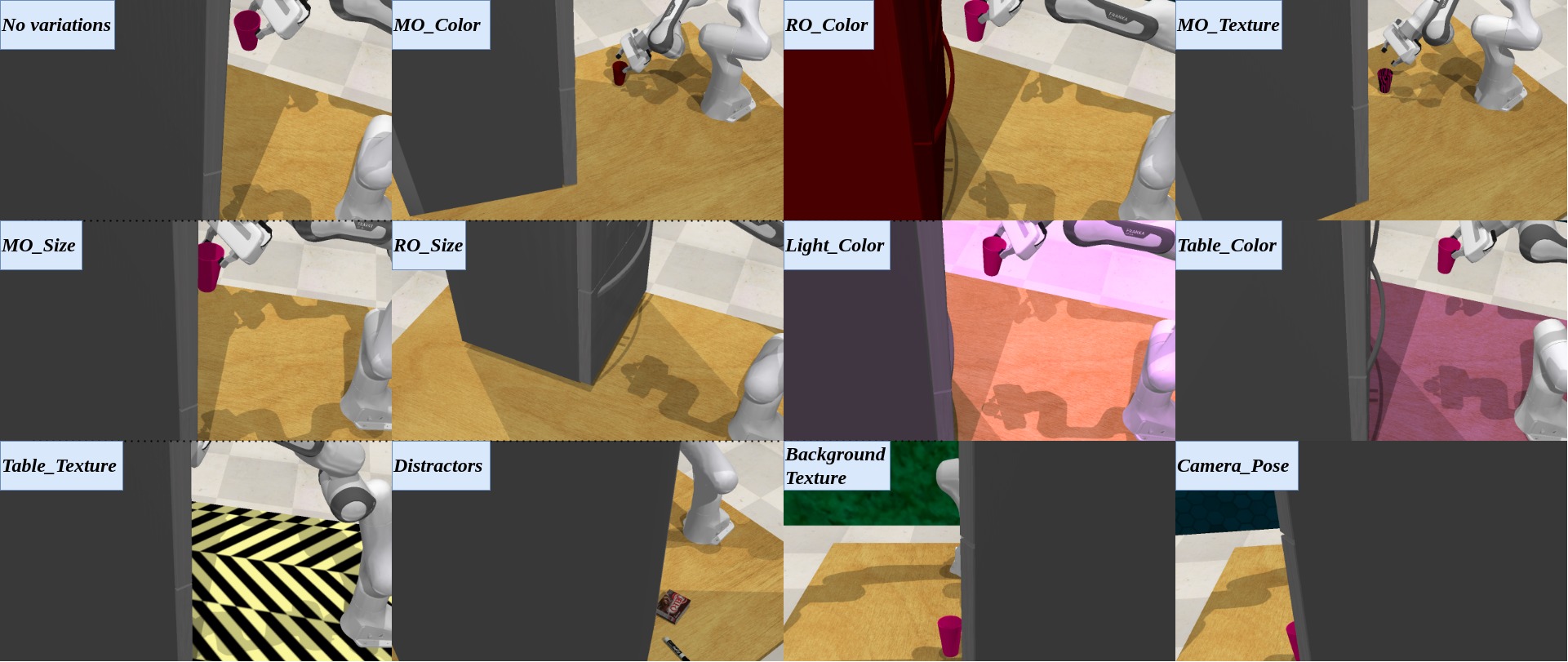}
    \caption{Perturbations for the \texttt{get\_ice\_from\_fridge} task}
    \label{fig:app_perturbations_get_ice_from_fridge}
\end{figure*}

\begin{figure*}
    \centering
    \includegraphics[width=\linewidth]{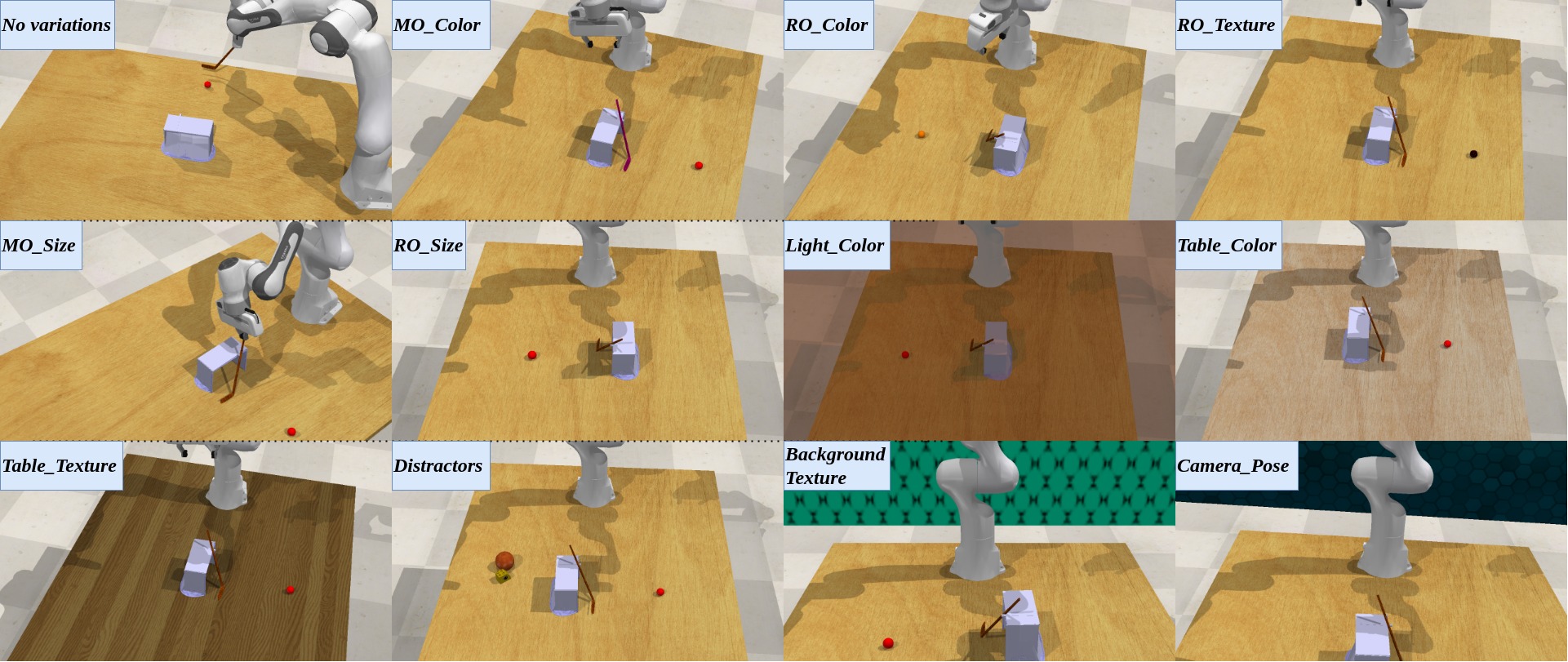}
    \caption{Perturbations for the \texttt{hockey} task}
    \label{fig:app_perturbations_hockey}
\end{figure*}

\begin{figure*}
    \centering
    \includegraphics[width=\linewidth]{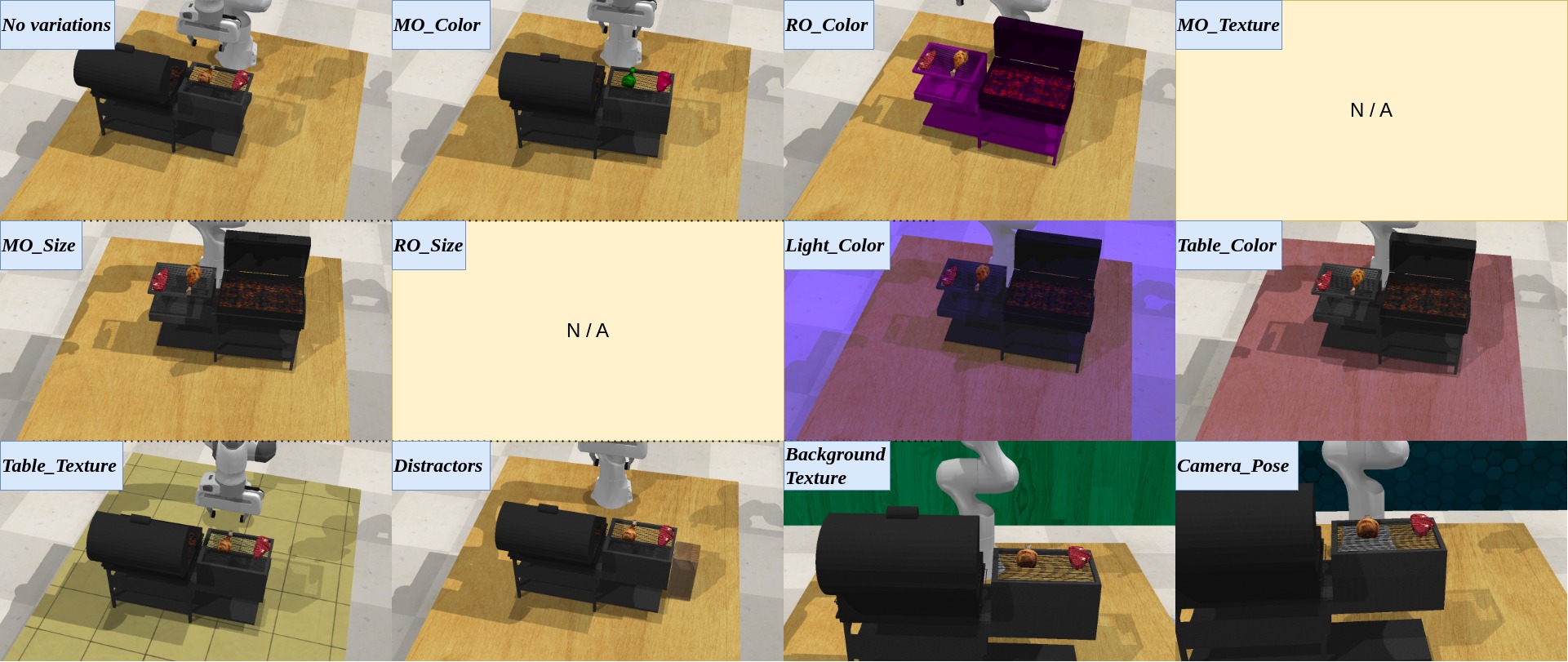}
    \caption{Perturbations for the \texttt{meat\_on\_grill} task}
    \label{fig:app_perturbations_meat_on_grill}
\end{figure*}

\begin{figure*}
    \centering
    \includegraphics[width=\linewidth]{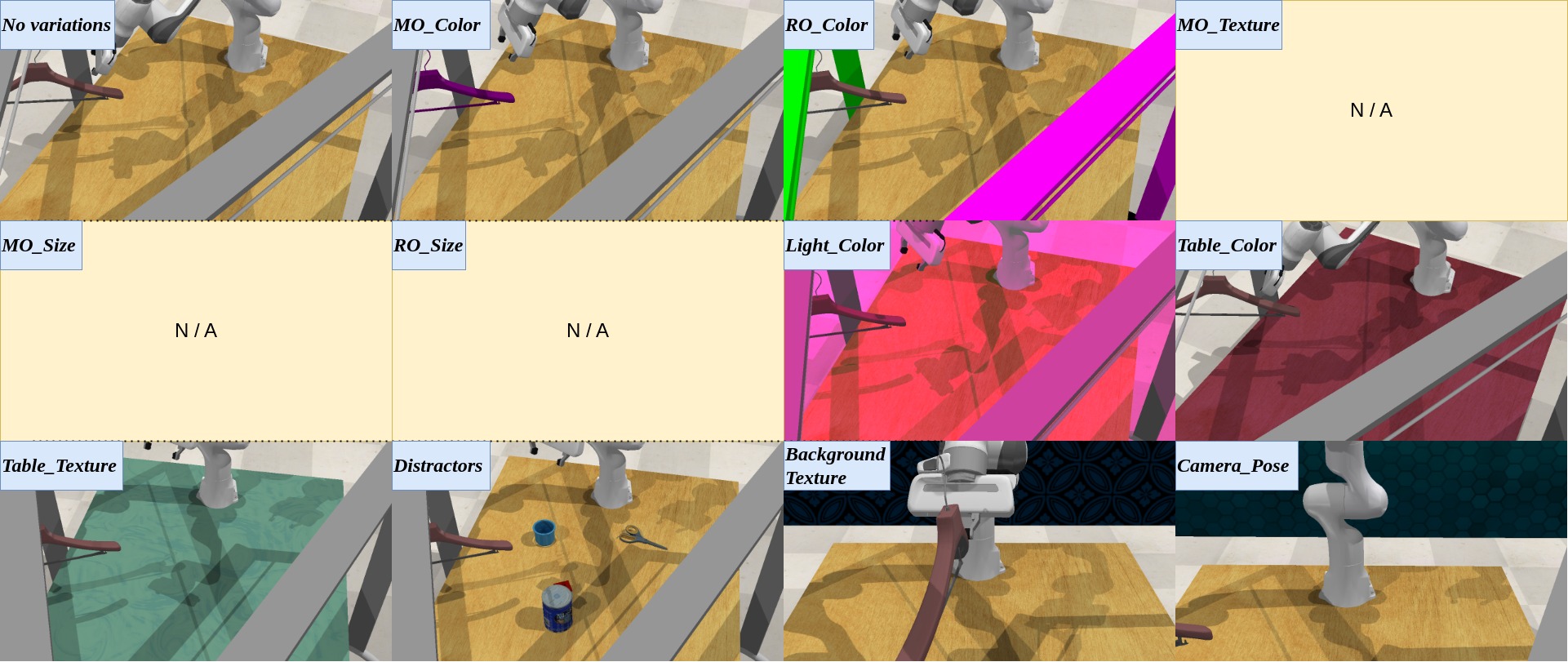}
    \caption{Perturbations for the \texttt{move\_hanger} task}
    \label{fig:app_perturbations_move_hanger}
\end{figure*}

\begin{figure*}
    \centering
    \includegraphics[width=\linewidth]{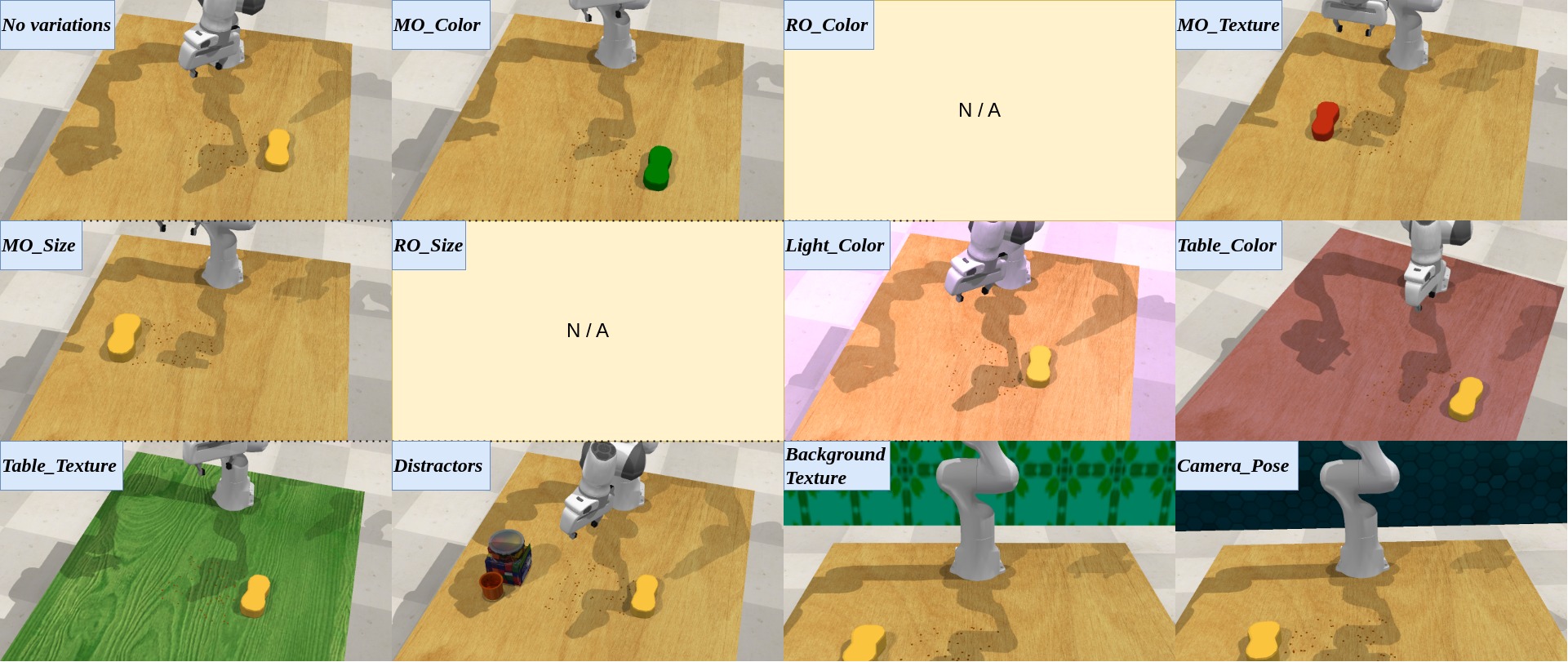}
    \caption{Perturbations for the \texttt{wipe\_desk} task}
    \label{fig:app_perturbations_wipe_desk}
\end{figure*}

\begin{figure*}
    \centering
    \includegraphics[width=\linewidth]{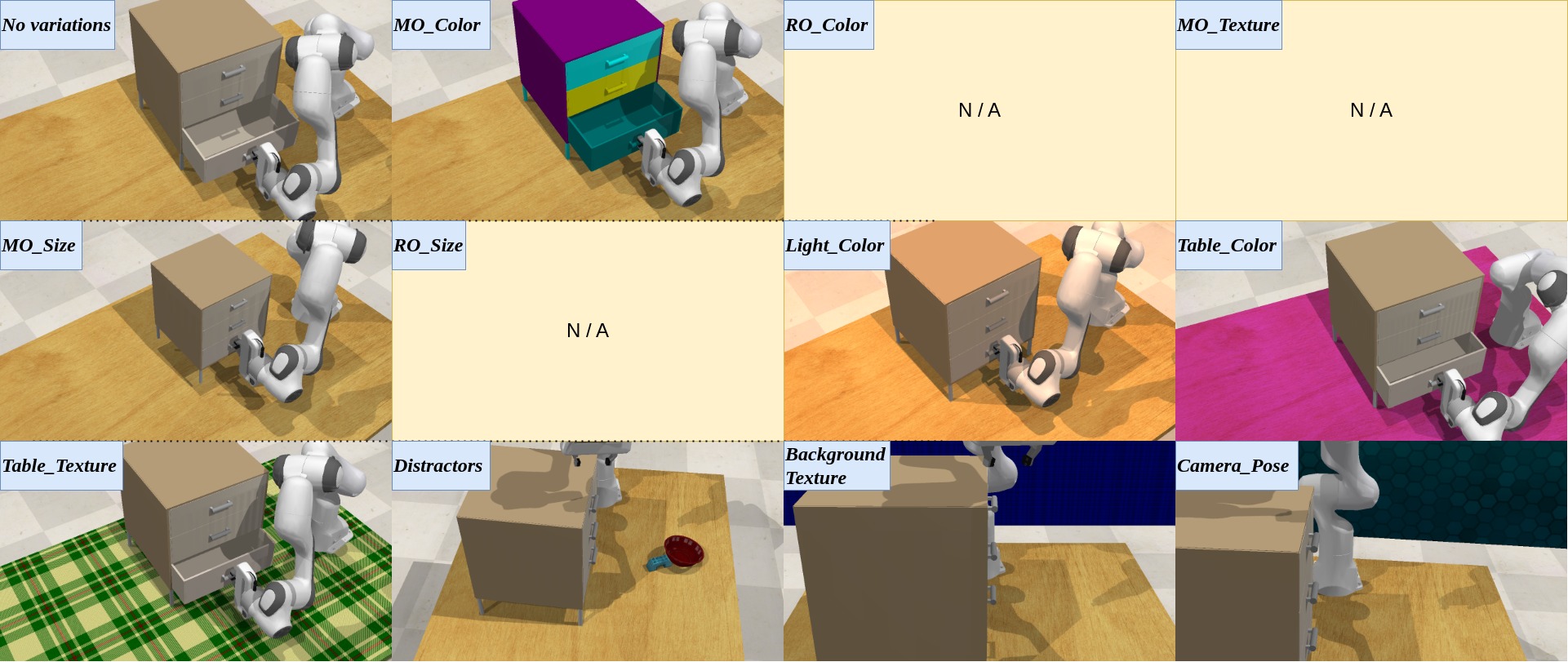}
    \caption{Perturbations for the \texttt{open\_drawer} task}
    \label{fig:app_perturbations_open_drawer}
\end{figure*}

\begin{figure*}
    \centering
    \includegraphics[width=\linewidth]{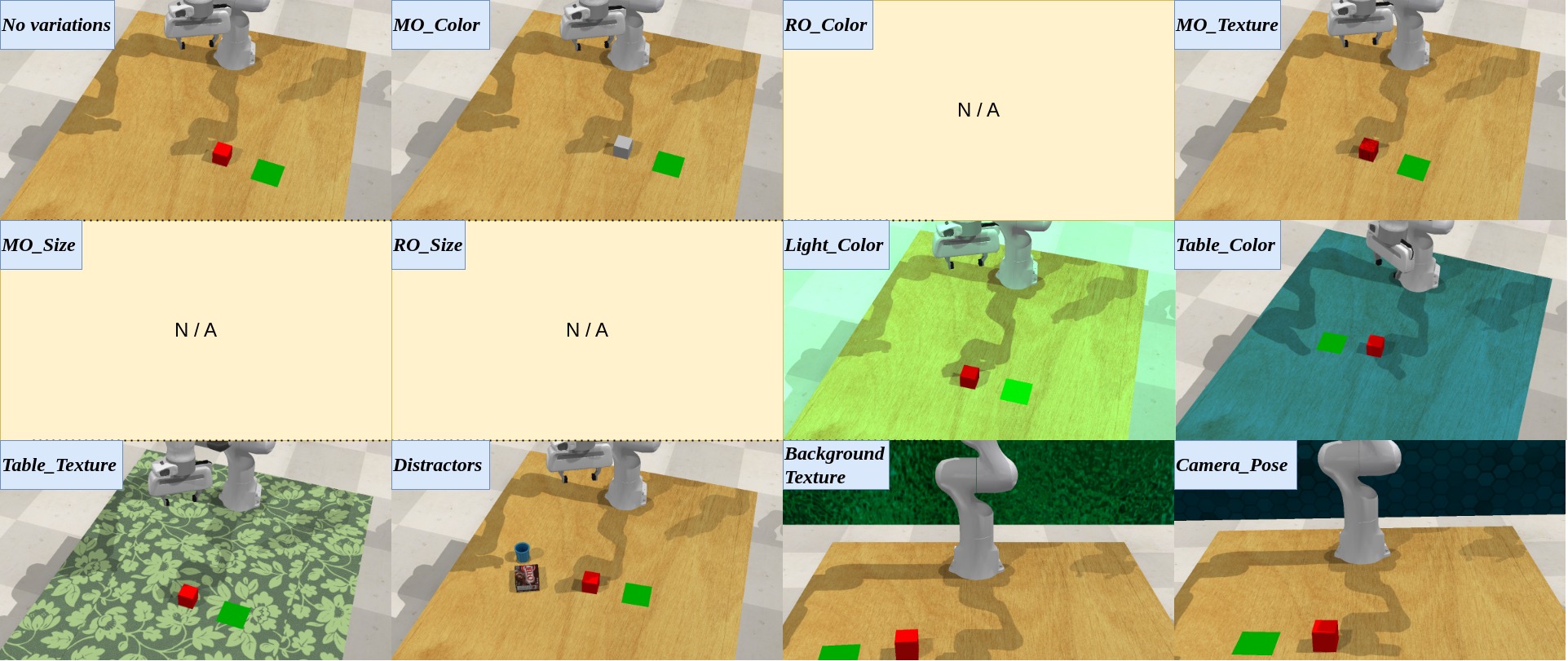}
    \caption{Perturbations for the \texttt{slide\_block\_to\_target} task}
    \label{fig:app_perturbations_slide_block_to_target}
\end{figure*}

\begin{figure*}
    \centering
    \includegraphics[width=\linewidth]{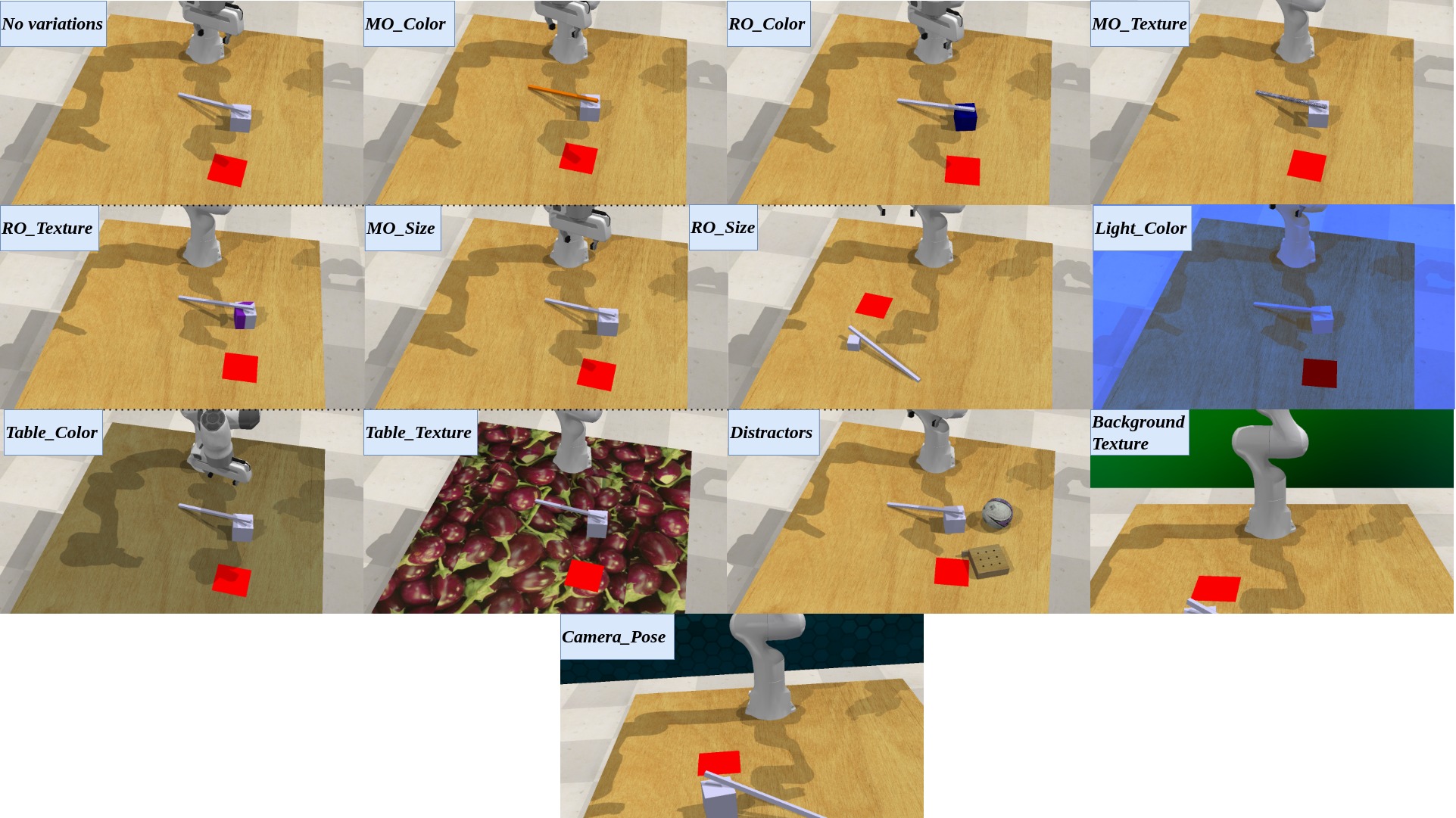}
    \caption{Perturbations for the \texttt{reach\_and\_drag} task}
    \label{fig:app_perturbations_reach_and_drag}
\end{figure*}

\begin{figure*}
    \centering
    \includegraphics[width=\linewidth]{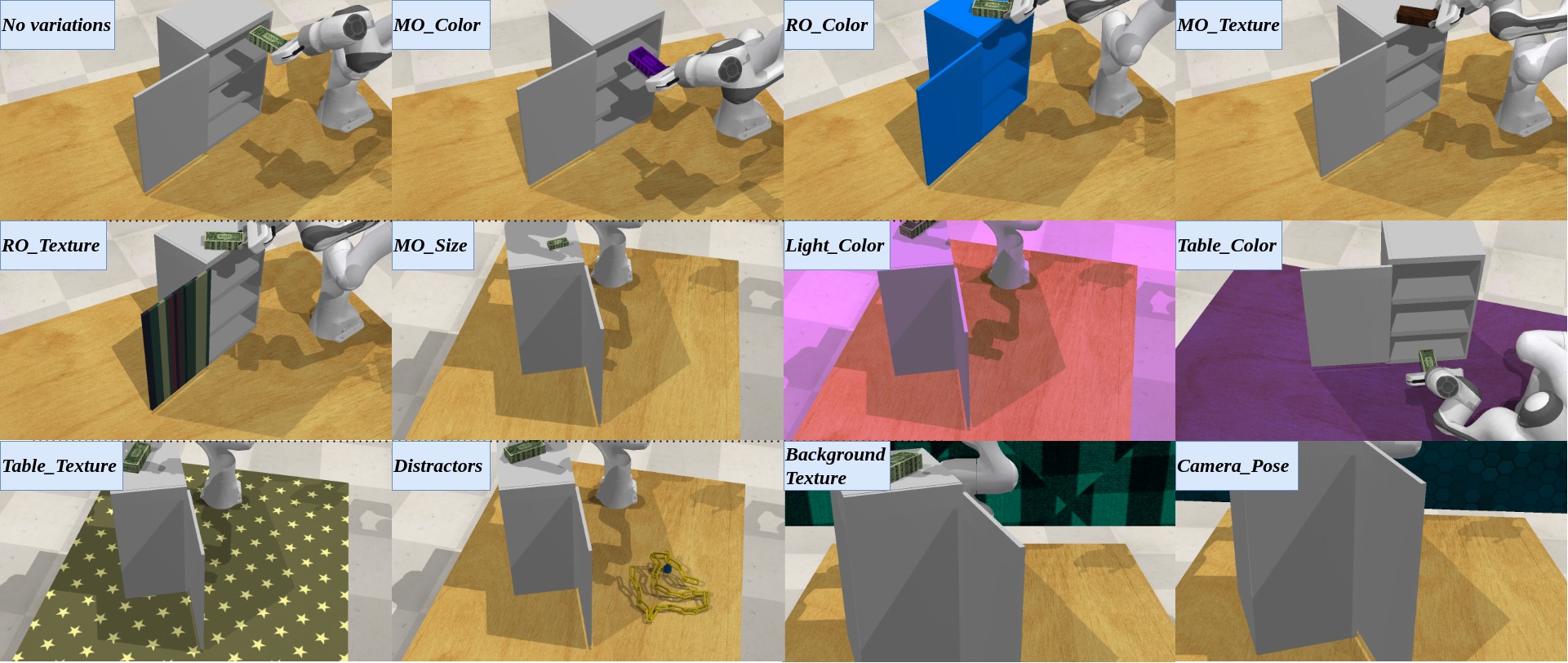}
    \caption{Perturbations for the \texttt{put\_money\_in\_safe} task}
    \label{fig:app_perturbations_put_money_in_safe}
\end{figure*}

\begin{figure*}
    \centering
    \includegraphics[width=\linewidth]{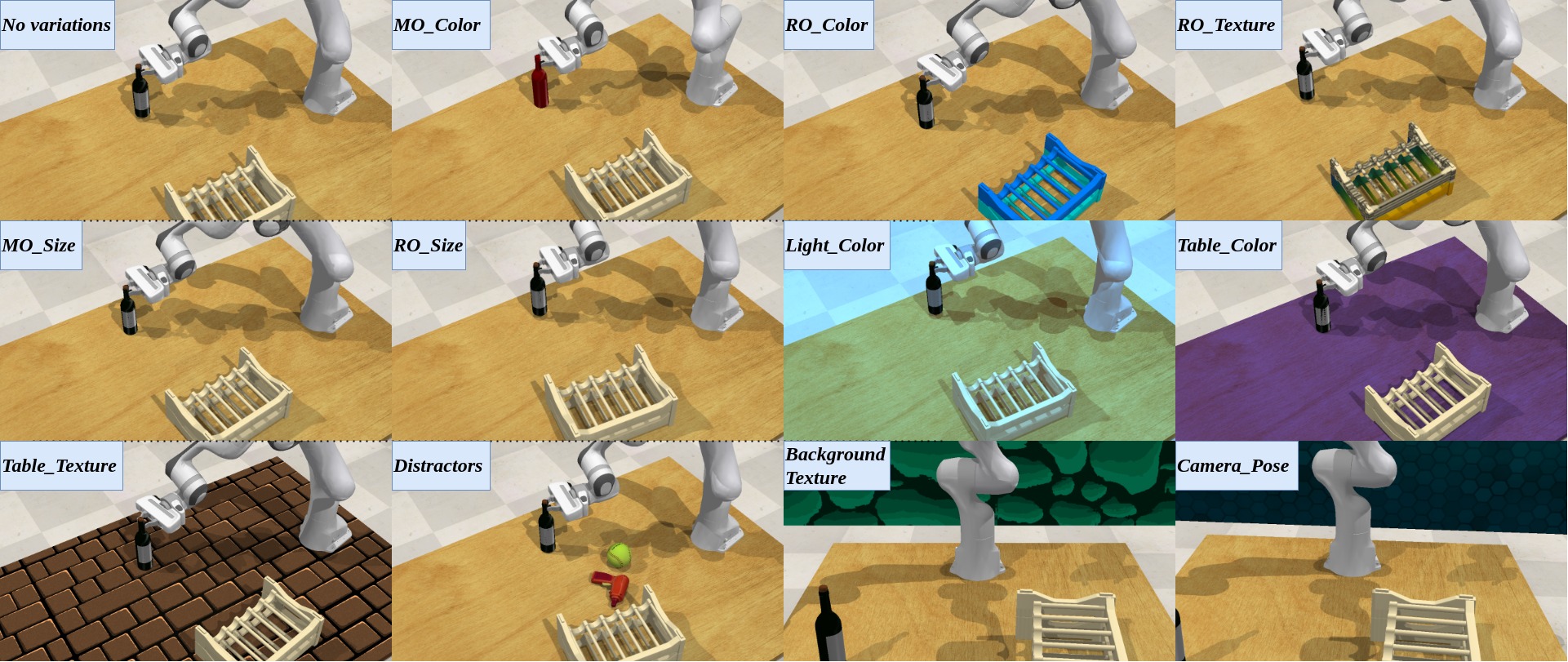}
    \caption{Perturbations for the \texttt{place\_wine\_at\_rack\_location} task}
    \label{fig:app_perturbations_place_wine_at_rack_location}
\end{figure*}

\begin{figure*}
    \centering
    \includegraphics[width=\linewidth]{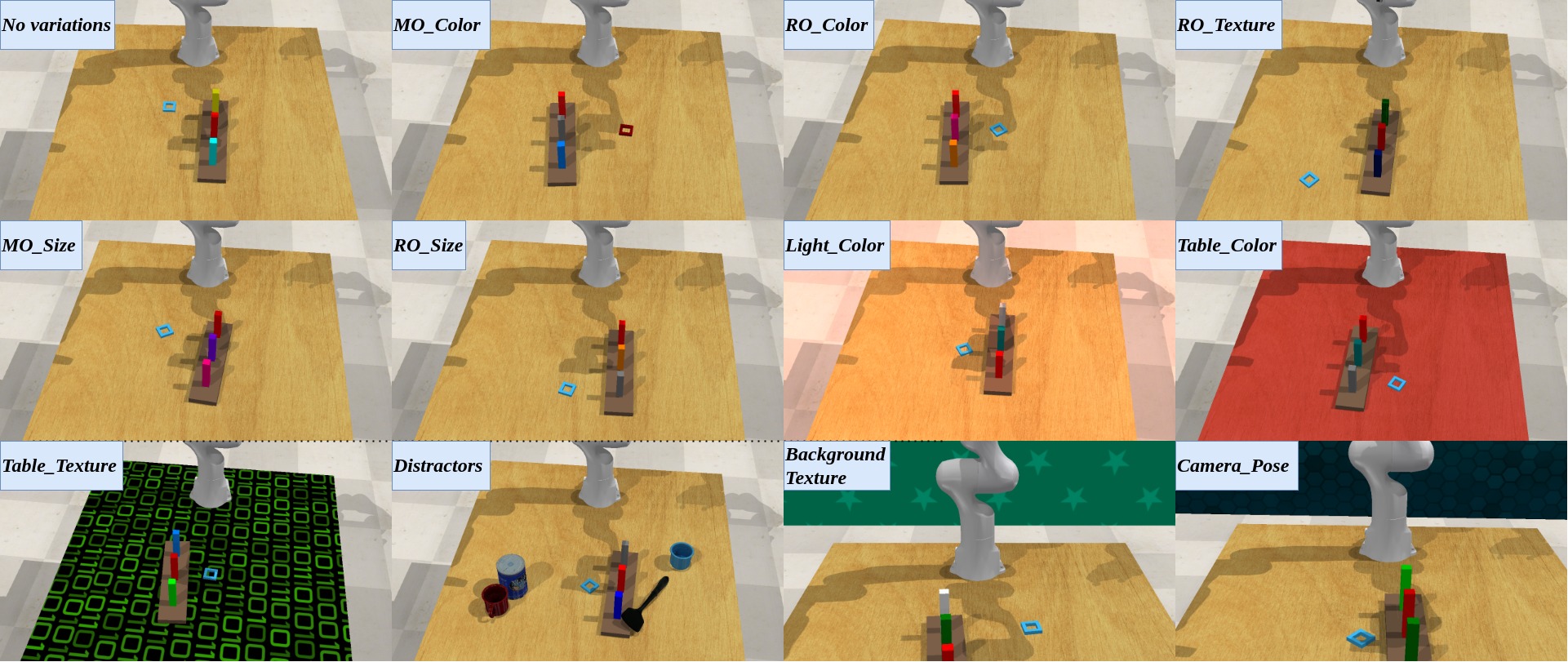}
    \caption{Perturbations for the \texttt{insert\_onto\_square\_peg\_location} task}
    \label{fig:app_perturbations_insert_onto_square_peg}
\end{figure*}

\begin{figure*}
    \centering
    \includegraphics[width=\linewidth]{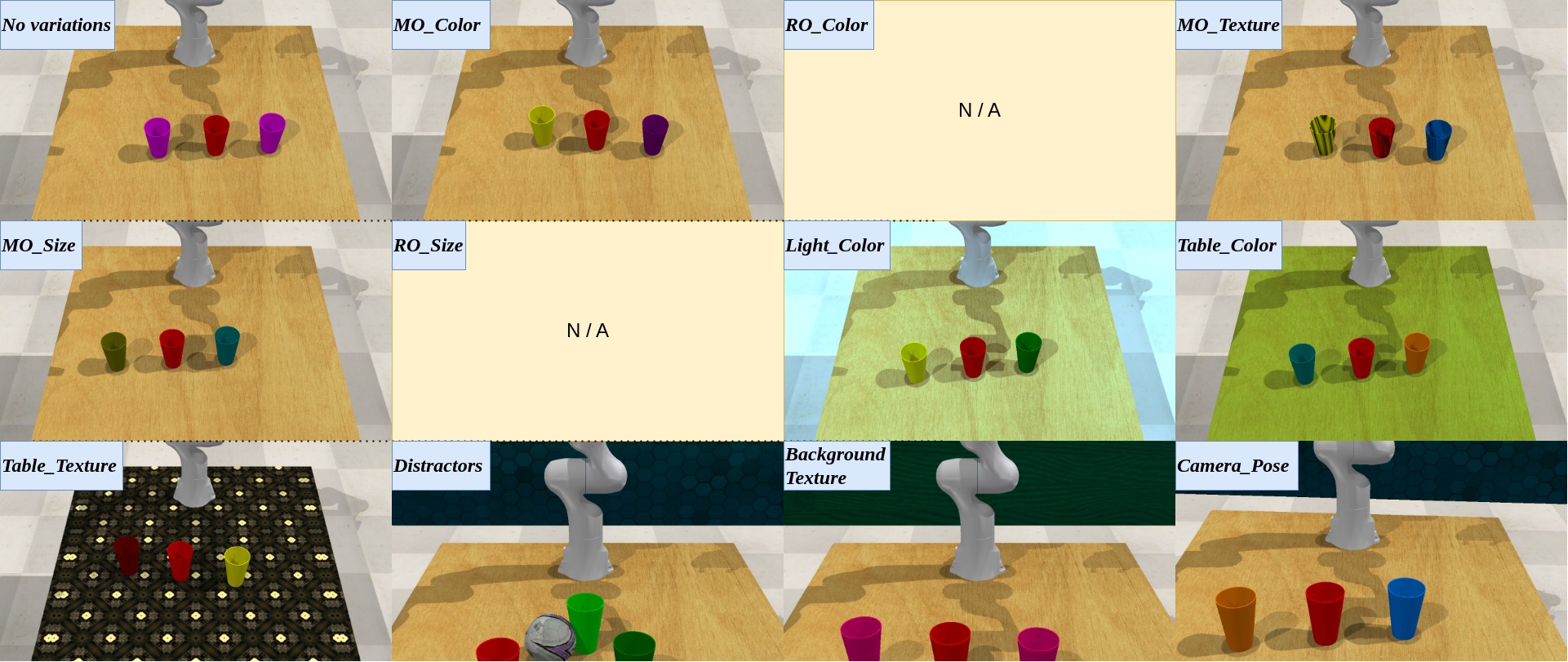}
    \caption{Perturbations for the \texttt{stack\_cups} task}
    \label{fig:app_perturbations_stack_cups}
\end{figure*}

\begin{figure*}
    \centering
    \includegraphics[width=\linewidth]{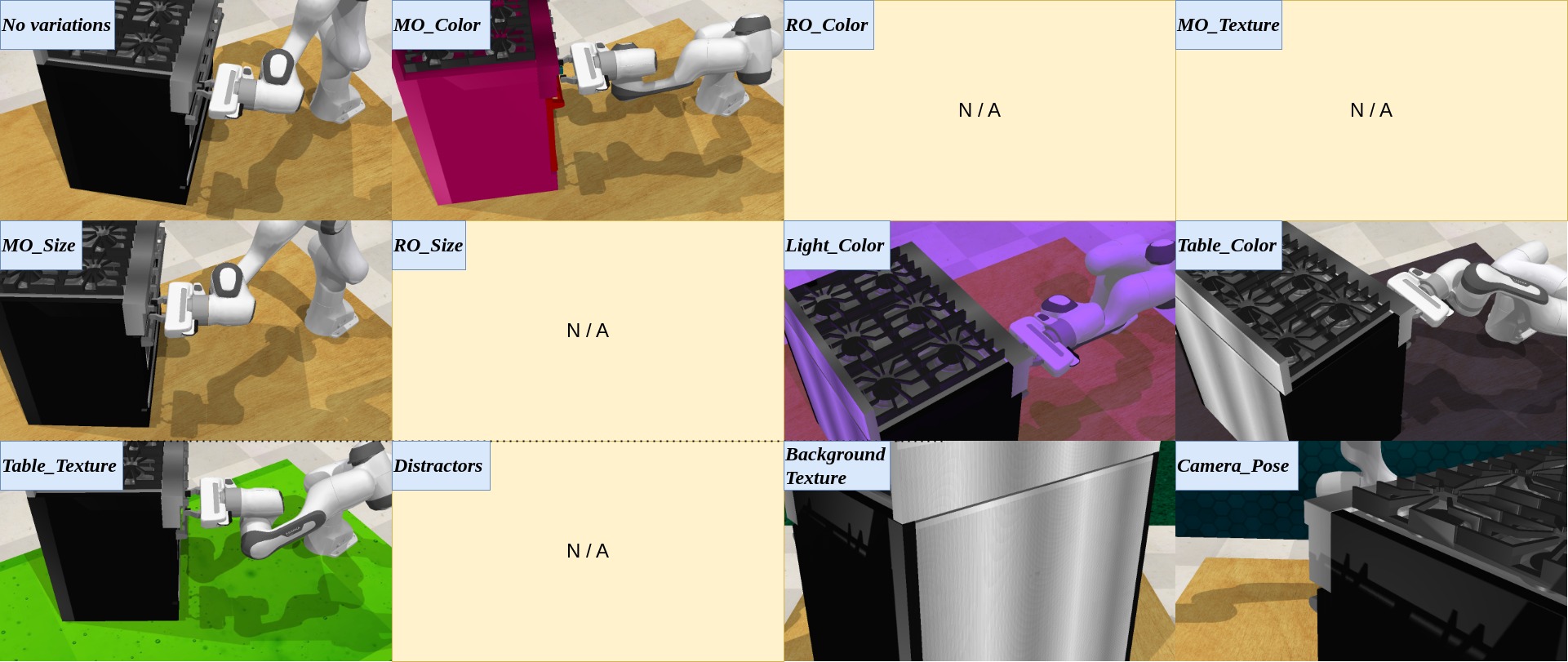}
    \caption{Perturbations for the \texttt{turn\_oven\_on} task}
    \label{fig:app_perturbations_turn_oven_on}
\end{figure*}

\begin{figure*}
    \centering
    \includegraphics[width=\linewidth]{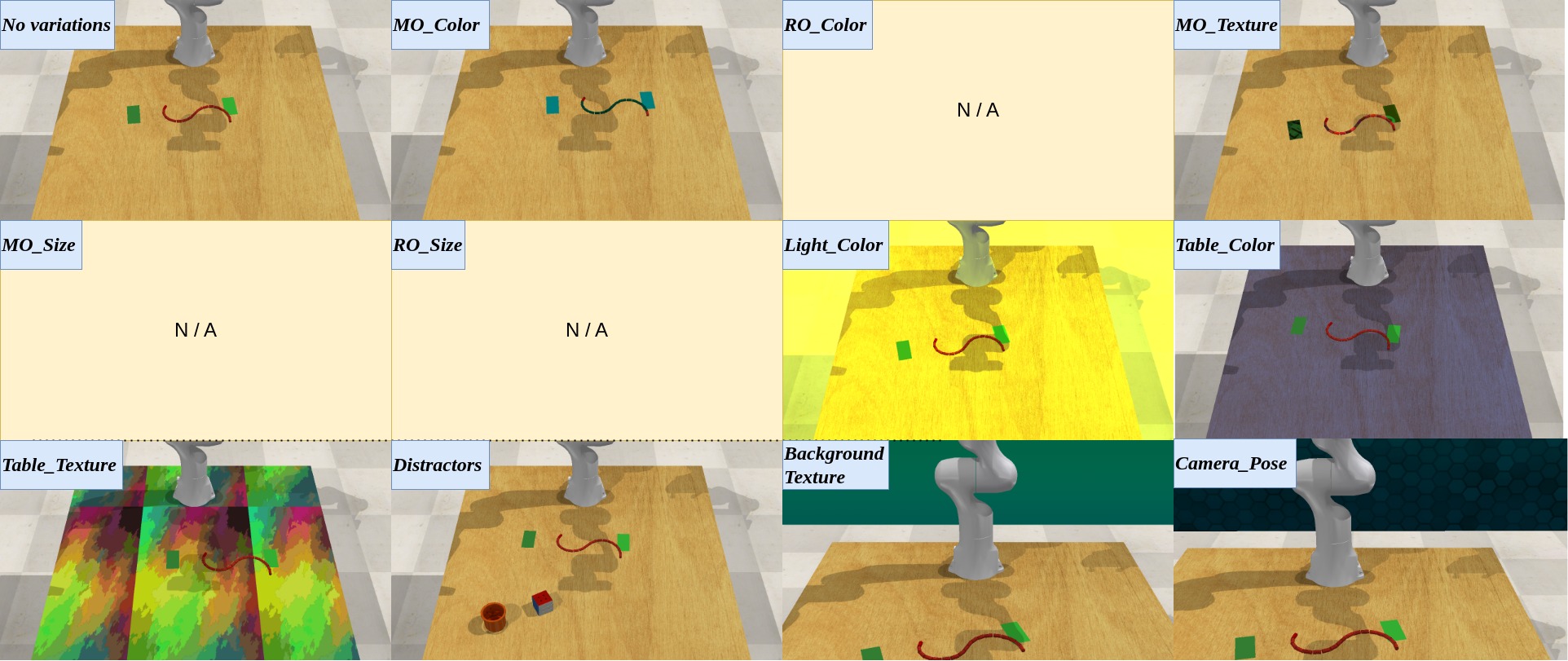}
    \caption{Perturbations for the \texttt{straighten\_rope} task}
    \label{fig:app_perturbations_straighten_rope}
\end{figure*}

\begin{figure*}
    \centering
    \includegraphics[width=\linewidth]{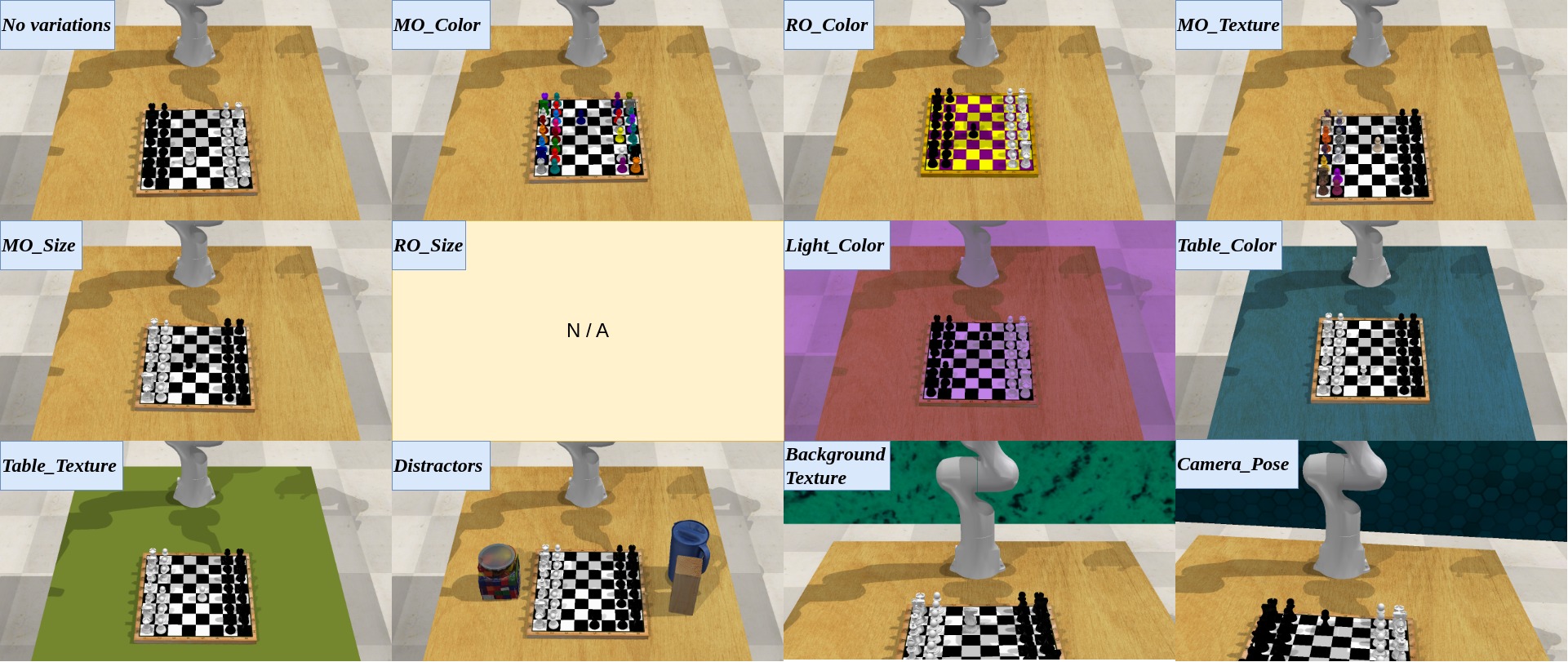}
    \caption{Perturbations for the \texttt{setup\_chess} task}
    \label{fig:app_perturbations_setup_chess}
\end{figure*}

\begin{figure*}
    \centering
    \includegraphics[width=\linewidth]{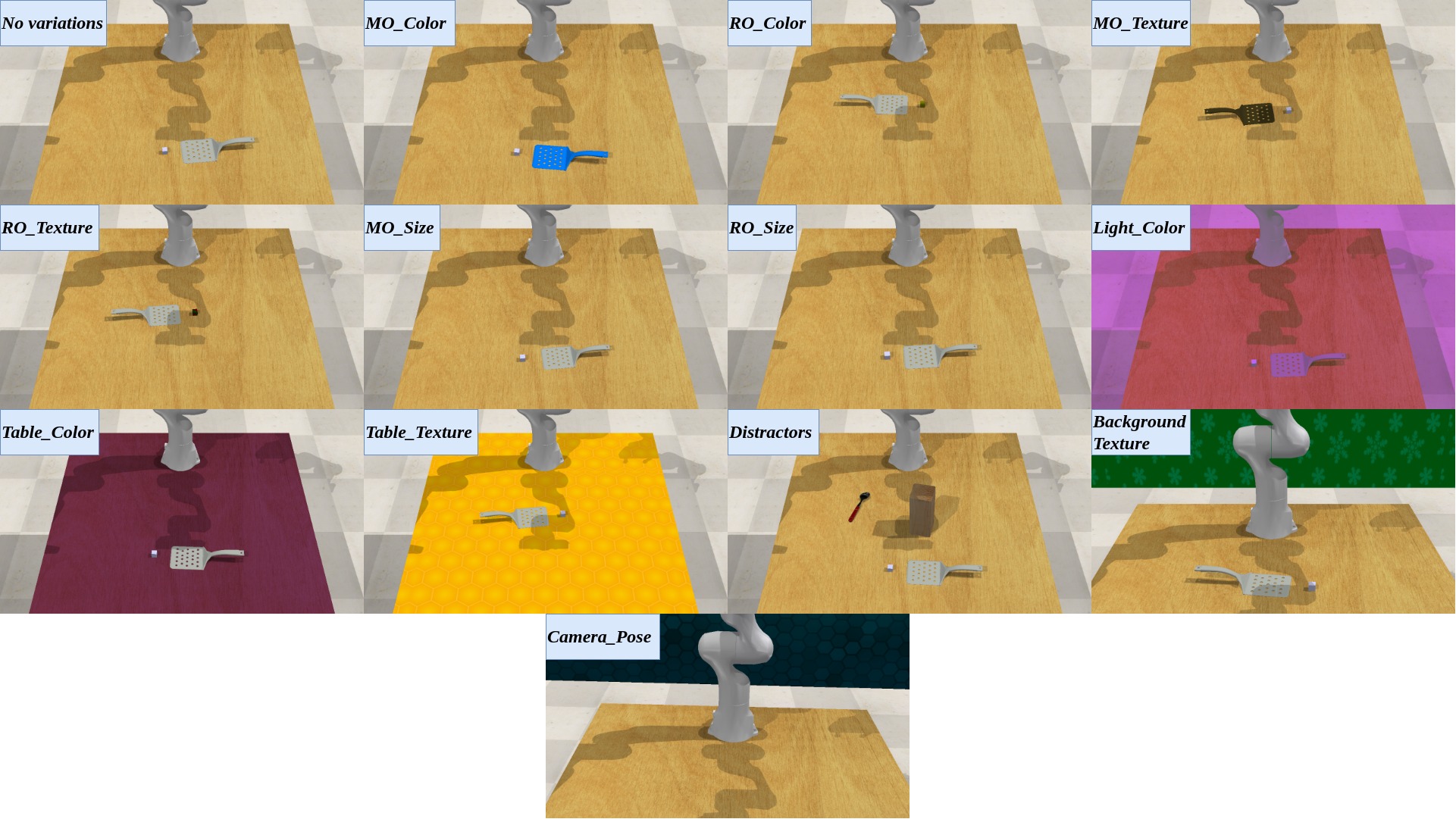}
    \caption{Perturbations for the \texttt{scoop\_with\_spatula} task}
    \label{fig:app_perturbations_scoop_with_spatula}
\end{figure*}

We provide full benchmark perturbation details for each task in Tables~\ref{tab:summary_of_variations_pt1}. Tables~\ref{tab:summary_of_variations_pt1} defines \texttt{MO} and \texttt{RO} objects for each task, specifies whether the applied perturbation is sampled from a discrete set or a continuous range, and finally provides the corresponding set size or the range. `-` means the perturbation does not apply due to either absence of \texttt{RO} for the task, or the simulator doesn't support that factor for the specified object. The remaining 6 \variationfactors\ apply to all the tasks. We specify their corresponding perturbation parameters in the main text (Section III.C). In Figure~\ref{fig:app_config_file}, we show an example of a task configuration file, and how its \variationfactors\ and their parameters can be specified or changed. In Figures~\ref{fig:app_perturbations_basketball}-~\ref{fig:app_perturbations_scoop_with_spatula}, we show all perturbed views for each task.

\subsection{Training details and Detailed results}
To train the baseline models, we use 1-4 NVIDIA RTX A6000 for 1-6 days. For a full evaluation over \benchname, we run multiple parallel jobs with batches launching in a sequence. Total compute used for this process was 4 NVIDIA RTX A6000 over 2-3 days for each model.

We report detailed per task success rates on each of the \variationfactors\ in Tables~\ref{tab:results_peract}-\ref{tab:results_rvt} for all the baselines.

\section{Real world details}
\subsection{Robot hardware setup}
\begin{figure*}[t]
  \centering
  \includegraphics[width=.6\linewidth]{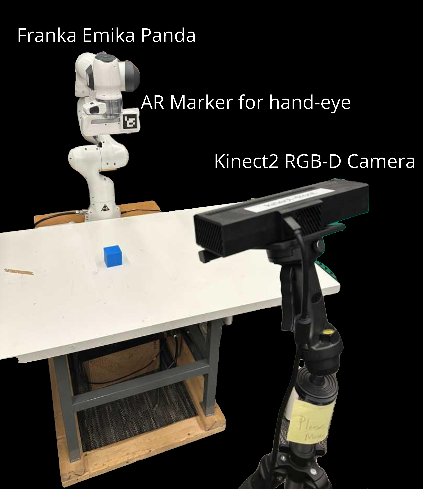}
  \caption{\textbf{Real-Robot Setup with Kinect-2 and Franka Panda.}}
  \label{fig:robtset}
\end{figure*}

The real-robot experiments use a Franka Panda manipulator with a parallel gripper. For perception, we use a Kinect-2 RGB-D camera mounted on a tripod, at an angle, pointing towards the tabletop.  Kinect-2 provides RGB-D images of resolution 512 × 424 at 30Hz. The extrinsic
between the camera and robot base-frame are calibrated with the easy hand-eye package. We use an ARUCO AR marker mounted on the gripper to aid the calibration process, as shown in Figure \ref{fig:robtset}.

\subsection{Task setup}
\begin{figure*}[t]
  \centering
  \includegraphics[width=\linewidth]{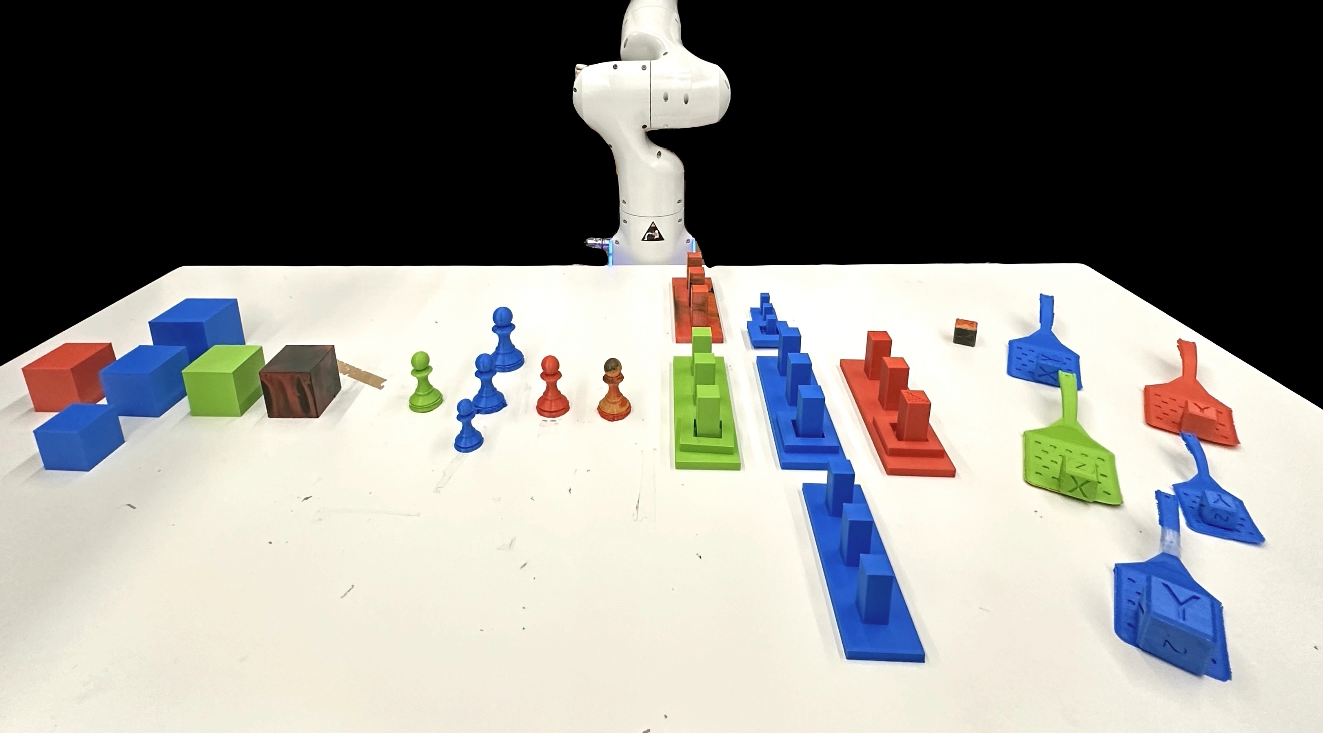}
  \caption{\textbf{3D print-outs of all the assets for the real-world tasks.}}
  \label{fig:3ds}
\end{figure*}

For the object assets, we 3D printed all of them as shown in Figure \ref{fig:3ds}. For (\texttt{RO}/\texttt{MO-Sizes}), we vary the scale by ±0.2 times original object size. For \texttt{RO}/\texttt{MO-Colors}, we use two different printing filaments (red and blue). For the \texttt{Light-Color} variation, we use a single color-changing spotlight. The success condition of each of the tasks are defined as follows:
\begin{itemize}
  \item \texttt{slide block to target}: Push the colored block into the light yellow patch with the word `target' written on it.
  \item \texttt{setup chess}: Pick up the pawn piece and put it onto the blue marked chess spot.
  \item \texttt{insert on square peg}: Pick up the colored square peg and insert it onto the right most pole.
  \item \texttt{scoop with spatula}: Push the spatula inwards to scoop up the cube, and then lift it up.
\end{itemize}

\subsection{Data collection}
We gather data through demonstrations using an HTC Vive controller, a device capable of 6 degrees of freedom (DoF) tracking, ensuring precise positioning relative to a stationary base station. The positions captured are visualized in RViz as markers on the real-time RGB-D pointcloud data obtained from the Kinect camera sensor. Users determine desired positions to record as keypoints by referencing both the marker and the pointcloud. These specified positions are then realized through the employment of a motion planning algorithm. For this purpose, we employ the Franka ROS interface along with MoveIt, which inherently utilizes the RRT-Connect planning algorithm by default.

\subsection{Training and Evaluation details}
The real robot's training was run on 1 NVIDIA TITAN RTX GPU for 1 day. 
We monitor the keypoints predicted by the real-world model to verify the safety of the next action. The robot continues to execute predicted keypoints during evaluation unless manually halted by the human operator.

\begin{table*}[ht]
\centering 
\caption{Real-world ablation study} 
\label{tab:sample_table} 
\begin{tabular}{|c|c|c|c|} 
\hline 
\textbf{Combination of perturbations} & \textbf{\benchmark\'s perturbations} & \textbf{Realistic real-world scenarios} & \textbf{Correlation} \\ 
\hline
Distractors + MO\_Size & [30,30,20,30,10] & [10,0,10,10,30] & 0.75 \\ 
\hline
Distractors+ Light\_Color & [70,60,70,70,50], Cell 2 & [50,40,50,40,50] & 0.01\\  
\hline
Light\_Color+Table\_Texture+Distractor+MO\_Size & [30,40,40,40,30] & [30,10,10,20,30]& 0.83 \\ 
\hline
\end{tabular}
\end{table*}

\subsection{Ablation study}
To investigate the compound effects of multiple perturbations on model performance and their correlation with real-world scenarios, we conducted an ablation study using the task \texttt{slide\_block\_to\_target}. We selected three perturbation combinations from real-world experiments and constructed three analogous real-world scenarios: a workbench, a dining table, and a study room table. Each scenario was subjected to the same perturbations derived from the benchmark's combinations. We assessed \texttt{PerAct}, a model trained on these real-world experiments, across both sets of scenarios. Each scenario underwent 10 episodes across five trials. Our analysis revealed a strong correlation between two of the three scenarios, as detailed in the depicted in Table \ref{tab:sample_table}.

\begin{table*}
  \centering
  \resizebox{\textwidth}{!}{
  \begin{tabular}{lccccccccc}
    \hline
    \diaghead{\theadfont Task Variation}%
    {Task}{Variation}                         &    MO           &    RO   &  MO\_Color  &  MO\_Size       &  MO\_Texture  &  RO\_Color  &  RO\_Size       &  RO\_Texture   &  Object Mass   \\ \hline
                                              &     -           &     -   &   discrete  & continuous      &    discrete   &   discrete  & continuous      &    discrete    &   continuous   \\ \hline
    \texttt{basketball\_in\_hoop}             &   ball          &  hoop   &  \numcolors &  $[0.75, 1.25]$ &  \numtextures &  \numcolors &  $[0.75, 1.15]$ &        -       &        -       \\
    \texttt{close\_box}                       &   box           &    -    &  \numcolors &  $[0.75, 1.15]$ &       -       &       -     &        -        &        -       &        -       \\
    \texttt{close\_laptop\_lid}               &  laptop         &    -    &  \numcolors &  $[0.75, 1.00]$ &       -       &       -     &        -        &        -       &        -       \\
    \texttt{empty\_dishwasher}                &  dishwasher     &  plate  &  \numcolors &  $[0.80, 1.00]$ &       -       &  \numcolors &  $[0.80, 1.00]$ &  \numtextures  &        -       \\
    \texttt{get\_ice\_from\_fridge}           &    cup          &  fridge &  \numcolors &  $[0.75, 1.25]$ &  \numtextures &  \numcolors &  $[0.75, 1.00]$ &        -       &        -       \\
    \texttt{hockey}                           &    stick        &   ball  &  \numcolors &  $[0.95, 1.05]$ &       -       &  \numcolors &  $[0.75, 1.25]$ &  \numtextures  &  $[0.1, 0.5]$  \\
    \texttt{meat\_on\_grill}                  &   meat          &  grill  &  \numcolors &  $[0.65, 1.15]$ &       -       &  \numcolors &        -        &        -       &        -       \\
    \texttt{move\_hanger}                     &   hanger        &  pole   &  \numcolors &        -        &       -       &  \numcolors &        -        &        -       &        -       \\
    \texttt{wipe\_desk}                       &   sponge        &  beans  &  \numcolors &  $[0.75, 1.25]$ &  \numtextures &  \numcolors &        -        &        -       &  $[1.0, 5.0]$  \\
    \texttt{open\_drawer}                     &   drawer        &    -    &  \numcolors &  $[0.75, 1.00]$ &       -       &       -     &        -        &        -       &        -       \\
    \texttt{slide\_block\_to\_target}         &   block         &    -    &  \numcolors &        -        &  \numtextures &       -     &        -        &        -       &  $[1.0, 15.0]$ \\
    \texttt{reach\_and\_drag}                 &   stick         &  block  &  \numcolors &  $[0.80, 1.10]$ &  \numtextures &  \numcolors &  $[0.50, 1.00]$ &  \numtextures  &  $[0.5, 2.5]$  \\
    \texttt{put\_money\_in\_safe}             &   money         &  safe   &  \numcolors &  $[0.50, 1.00]$ &  \numtextures &  \numcolors &        -        &  \numtextures  &        -       \\
    \texttt{place\_wine\_at\_rack\_location}  &   bottle        &  shelve &  \numcolors &  $[0.85, 1.15]$ &       -       &  \numcolors &  $[0.85, 1.15]$ &  \numtextures  &        -       \\
    \texttt{insert\_onto\_square\_peg}        &   peg           &  spokes &  \numcolors &  $[1.00, 1.50]$ &       -       &  \numcolors &  $[0.85, 1.15]$ &  \numtextures  &        -       \\
    \texttt{stack\_cups}                      &   cups          &    -    &  \numcolors &  $[0.75, 1.25]$ &  \numtextures &       -     &        -        &        -       &        -       \\
    \texttt{turn\_oven\_on}                   &   knobs         &    -    &  \numcolors &  $[0.50, 1.50]$ &       -       &       -     &        -        &        -       &        -       \\
    \texttt{straighten\_rope}                 &   rope          &    -    &  \numcolors &        -        &  \numtextures &       -     &        -        &        -       &        -       \\
    \texttt{setup\_chess}                     &   chess pieces  &  board  &  \numcolors &  $[0.75, 1.25]$ &  \numtextures &  \numcolors &        -        &        -       &        -       \\
    \texttt{scoop\_with\_spatula}             &   spatula       &  block  &  \numcolors &  $[0.75, 1.25]$ &  \numtextures &  \numcolors &  $[0.75, 1.50]$ &  \numtextures  &  $[1.0, 5.0]$  \\ \hline
  \end{tabular}
  }
  \caption{Summary of tasks and their \variationfactors. The table specifies when a certain factor is applied to a certain task and its corresponding parameters.}
  \label{tab:summary_of_variations_pt1}
\end{table*}

\begin{table*}[]
    \centering
    \begin{tabular}{lccccccccccccccccc}
    Task Name                       & \myrotcell{No variations} & \myrotcell{All variations} & \myrotcell{MO\_Color} & \myrotcell{RO\_Color} & \myrotcell{MO\_Texture} & \myrotcell{RO\_Texture} & \myrotcell{MO\_Size} & \myrotcell{RO\_Size} & \myrotcell{Light-color} & \myrotcell{Table-Color} & \myrotcell{Table-texture} & \myrotcell{Distractor} & \myrotcell{Backgrond-texture} & \myrotcell{RLBench variations} & \myrotcell{Camera pose} & \myrotcell{Object Friction} & \myrotcell{Object Mass} \\ \hline
    basketball\_in\_hoop            & 100           & 0              & 100       & 72        & 100        & -          & 100     & 74      & 92          & 84          & 76            & 48         & 96                & 100                & 96            & -             & -           \\
    close\_box                      & 65            & 0              & 28        & -         & -          & -          & 40      & -       & 50          & 15          & 28            & 30         & 50                & 52                 & 64            & -             & -           \\
    close\_laptop\_lid              & 96            & 80             & 80        & -         & -          & -          & 100     & -       & 92          & 88          & 80            & 88         & 100               & 100                & 96            & -             & -           \\
    empty\_dishwasher               & 0             & 0              & 0         & 0         & -          & 0          & 0       & 0       & 0           & 0           & 0             & 0          & 0                 & 4                  & 0             & -             & -           \\
    get\_ice\_from\_fridge          & 60            & 4              & 60        & 60        & 56         & -          & 60      & 68      & 60          & 76          & 40            & 72         & 76                & 76                 & 84            & -             & -           \\
    hockey                          & 0             & 0              & 0         & 0         & -`         & 0          & 0       & 0       & 0           & 0           & 0             & 0          & 0                 & 0                  & 0             & 0             & 0           \\
    meat\_on\_grill                 & 92            & 44             & 64        & 72        & -          & -          & 92      & -       & 64          & 92          & 60            & 88         & 80                & 92                 & 84            & -             & -           \\
    move\_hanger                    & 0             & 0              & 0         & 0         & -          & -          & -       & -       & 0           & 0           & 0             & 0          & 0                 & 0                  & 0             & -             & -           \\
    wipe\_desk                      & 0             & 0              & 0         & 0         & 0          & -          & 0       & -       & 0           & 0           & 0             & 0          & 0                 & 0                  & 0             & 0             & 0           \\
    open\_drawer                    & 28            & 0              & 0         & -         & -          & -          & 0       & -       & 16          & 80          & 32            & 28         & 8                 & 68                 & 76            & -             & -           \\
    slide\_block\_to\_target        & 24            & 0              & 4         & -         & 16         & -          & 68      & -       & 20          & 8           & 4             & 12         & 32                & 32                 & 0             & -             & -           \\
    reach\_and\_drag                & 36            & 0              & 20        & 12        & 4          & 8          & 40      & 8       & 12          & 12          & 8             & 0          & 20                & 64                 & 20            & 12            & 24          \\
    put\_money\_in\_safe            & 32            & 0              & 32        & 16        & 44         & 28         & 20      & -       & 28          & 12          & 12            & 20         & 20                & 44                 & 20            & -             & -           \\
    place\_wine\_at\_rack\_location & 0             & 0              & 0         & 0         & -          & 0          & 8       & 12      & 8           & 0           & 4             & 0          & 4                 & 8                  & 8             & -             & -           \\
    insert\_onto\_square\_peg       & 4             & 0              & 0         & 4         & -          & 4          & 0       & 8       & 8           & 4           & 0             & 8          & 4                 & 28                 & 0             & -             & -           \\
    stack\_cups                     & 8             & 0              & 12        & -         & 0          & -          & 0       & -       & 0           & 16          & 0             & 0          & 4                 & 0                  & 8             & -             & -           \\
    turn\_oven\_on                  & 24            & 8              & 20        & -         & -          & -          & 40      & -       & 40          & 40          & 48            & 40         & 36                & 32                 & 40            & -             & -           \\
    straighten\_rope                & 0             & 0              & 0         & -         & 0          & -          & -       & -       & 0           & 0           & 0             & 0          & 0                 & 4                  & 14            & -             & -           \\
    setup\_chess                    & 44            & 8              & 28        & 76        & 44         & -          & 0       & -       & 56          & 64          & 64            & 48         & 68                & 16                 & 60            & -             & -           \\
    scoop\_with\_spatula            & 76            & 0              & 32        & 68        & 24         & 84         & 72      & 64      & 36          & 16          & 8             & 60         & 72                & 68                 & 56            & 64            & 64         
    \end{tabular}
    \caption{Results for \texttt{PerAct} for various perturbations}
    \label{tab:results_peract}
    \vspace{1cm}
\end{table*}

\begin{table*}[]
    \centering
    \begin{tabular}{lccccccccccccccccc}
    Task Name                       & \myrotcell{No variations} & \myrotcell{All variations} & \myrotcell{MO\_Color} & \myrotcell{RO\_Color} & \myrotcell{MO\_Texture} & \myrotcell{RO\_Texture} & \myrotcell{MO\_Size} & \myrotcell{RO\_Size} & \myrotcell{Light-color} & \myrotcell{Table-Color} & \myrotcell{Table-texture} & \myrotcell{Distractor} & \myrotcell{Background-texture} & \myrotcell{RLBench variations} & \myrotcell{Camera pose} & \myrotcell{Object Friction} & \myrotcell{Object Mass}  \\ \hline
    basketball\_in\_hoop            & 0             & 0              & 0         & 0         & 0          & -          & 0       & 0       & 0           & 0           & 0             & 0          & 0                  & 0                   &  0           & -            & -             \\
    close\_box                      & 32            & 0              & 0         & -         & -          & -          & 24      & -       & -           & 12          & 0             & 0          & 12                 & 24                  &  8           & -            & -             \\
    close\_laptop\_lid              & 6             & 4              & 4         & -         & -          & -          & 4       & -       & 8           & 12          & 0             & 20         & 4                  & 4                   &  4           & -            & -             \\
    empty\_dishwasher               & 0             & 0              & 0         & 0         & -          & 0          & 0       & 0       & 0           & 0           & 0             & 0          & 0                  & 0                   &  0           & -            & -             \\
    get\_ice\_from\_fridge          & 0             & 0              & 0         & 0         & 0          & -          & 0       & 0       & 0           & 0           & 0             & 0          & 0                  & 0                   &  0           & -            & -             \\
    hockey                          & 0             & 0              & 0         & 0         & -          & 0          & 0       & 0       & 0           & 0           & 0             & 0          & 0                  & 0                   &  0           & 0            & 0             \\
    meat\_on\_grill                 & 0             & 0              & 0         & 0         & -          & -          & 0       & -       & 0           & 0           & 0             & 0          & 0                  & 0                   &  0           & -            & -             \\
    move\_hanger                    & 4             & 0              & 0         & 0         & -          & -          & -       & -       & 0           & 0           & 0             & 0          & 0                  & 0                   &  0           & -            & -             \\
    wipe\_desk                      & 0             & 0              & 0         & 0         & 0          & -          & 0       & -       & 0           & 0           & 0             & 0          & 0                  & 0                   &  0           & 0            & 0             \\
    open\_drawer                    & 0             & 0              & 0         & -         & -          & -          & 0       & -       & 0           & 0           & 0             & 0          & 0                  & 0                   &  0           & -            & -             \\
    slide\_block\_to\_target        & 0             & 0              & 0         & -         & 0          & -          & 0       & -       & 0           & 0           & 0             & 0          & 0                  & 0                   &  0           & -            & -             \\
    reach\_and\_drag                & 0             & 0              & 0         & 0         & 0          & 0          & 0       & 0       & 0           & 0           & 0             & 0          & 0                  & 0                   &  0           & 0            & 0            \\
    put\_money\_in\_safe            & 0             & 0              & 0         & 0         & 0          & 0          & 0       & -       & 0           & 0           & 0             & 0          & 0                  & 0                   &  0           & -            & -             \\
    place\_wine\_at\_rack\_location & 0             & 0              & 0         & 0         & -          & 0          & 0       & 0       & 0           & 0           & 0             & 0          & 0                  & 0                   &  0           & -            & -             \\
    insert\_onto\_square\_peg       & 0             & 0              & 0         & 0         & -          & 0          & 0       & 0       & 0           & 0           & 0             & 0          & 0                  & 0                   &  0           & -            & -             \\
    stack\_cups                     & 0             & 0              & 0         & -         & 0          & -          & 0       & -       & 0           & 0           & 0             & 0          & 0                  & 0                   &  0           & -            & -             \\
    turn\_oven\_on                  & 12            & 8              & 4         & -         & -          & -          & 4       & -       & 12          & 4           & 4             & 12         & 8                  & 12                  &  4           & -            & -             \\
    straighten\_rope                & 0             & 0              & 0         & -         & 0          & -          & -       & -       & 0           & 0           & 0             & 0          & 0                  & 0                   &  0           & -            & -             \\
    setup\_chess                    & 4             & 0              & 0         & 0         & 0          & -          & 0       & -       & 0           & 0           & 0             & 0          & 0                  & 0                   &  0           & -            & -             \\
    scoop\_with\_spatula            & 0             & 0              & 0         & 0         & 0          & 0          & 0       & 0       & 0           & 0           & 0             & 0          & 0                  & 0                   &  0           & 0            & 0           
    \end{tabular}
    \caption{Results for \texttt{R3M} for various perturbations}
    \label{tab:results_r3m}
    \vspace{1cm}
\end{table*}

\clearpage

\begin{table*}[]
    \centering
    \begin{tabular}{lccccccccccccccccc}
    Task Name                       & \myrotcell{No variations} & \myrotcell{All variations} & \myrotcell{MO\_Color} & \myrotcell{RO\_Color} & \myrotcell{MO\_Texture} & \myrotcell{RO\_Texture} & \myrotcell{MO\_Size} & \myrotcell{RO\_Size} & \myrotcell{Light-color} & \myrotcell{Table-Color} & \myrotcell{Table-texture} & \myrotcell{Distractor} & \myrotcell{Background-texture} & \myrotcell{RLBench variations} & \myrotcell{Camera pose} & \myrotcell{Object Friction} & \myrotcell{Object Mass}  \\ \hline
    basketball\_in\_hoop            & 4             & 0              & 4         & 0         & 4          & -          & 0       & 4       & 4           & 4           & 0             & 4          & 0                  & 0                  & 0           & -            & -          \\
    close\_box                      & 40            & 8              & 12        & -         & -          & -          & 60      & -       & 8           & 8           & 12            & 8          & 36                 & 24                 & 12          & -            & -          \\
    close\_laptop\_lid              & 8             & 0              & 8         & -         & -          & -          & 0       & -       & 16          & 4           & 4             & 40         & 0                  & 4                  & 20          & -            & -          \\
    empty\_dishwasher               & 0             & 0              & 0         & 0         & -          & 0          & 0       & 0       & 0           & 0           & 0             & 0          & 0                  & 0                  & 0           & -            & -          \\
    get\_ice\_from\_fridge          & 0             & 0              & 0         & 0         & 0          & -          & 0       & 0       & 0           & 0           & 0             & 0          & 0                  & 0                  & 0           & -            & -          \\
    hockey                          & 0             & 0              & 0         & 0         & -          & 0          & 0       & 0       & 0           & 0           & 0             & 0          & 0                  & 0                  & 0           & 0            & 0          \\
    meat\_on\_grill                 & 10            & 4              & 0         & 0         & -          & -          & 0       & -       & 0           & 0           & 0             & 12         & 0                  & 0                  & 4           & -            & -          \\
    move\_hanger                    & 0             & 0              & 0         & 0         & -          & -          & -       & -       & 0           & 0           & 0             & 0          & 0                  & 0                  & 0           & -            & -          \\
    wipe\_desk                      & 0             & 0              & 0         & 0         & 0          & -          & 0       & -       & 0           & 0           & 0             & 0          & 0                  & 0                  & 0           & 0            & 0          \\
    open\_drawer                    & 0             & 0              & 0         & -         & -          & -          & 0       & -       & 0           & 0           & 0             & 0          & 0                  & 0                  & 0           & -            & -          \\
    slide\_block\_to\_target        & 0             & 0              & 0         & -         & 0          & -          & 0       & -       & 0           & 0           & 0             & 0          & 0                  & 0                  & 0           & -            & -          \\
    reach\_and\_drag                & 0             & 0              & 0         & 0         & 0          & 0          & 0       & 0       & 0           & 0           & 0             & 0          & 0                  & 0                  & 0           & 0            & 0          \\
    put\_money\_in\_safe            & 0             & 0              & 0         & 0         & 0          & 0          & 0       & -       & 0           & 0           & 0             & 0          & 0                  & 0                  & 0           & -            & -          \\
    place\_wine\_at\_rack\_location & 0             & 0              & 0         & 0         & -          & 0          & 0       & 0       & 0           & 0           & 0             & 0          & 0                  & 0                  & 0           & -            & -          \\
    insert\_onto\_square\_peg       & 0             & 0              & 0         & 0         & -          & 0          & 0       & 0       & 0           & 0           & 0             & 0          & 0                  & 0                  & 0           & -            & -          \\
    stack\_cups                     & 0             & 0              & 0         & -         & 0          & -          & 0       & -       & 0           & 0           & 0             & 0          & 0                  & 0                  & 0           & -            & -          \\
    turn\_oven\_on                  & 6             & 4              & 0         & -         & -          & -          & 20      & -       & 4           & 16          & 4             & 12         & 8                  & 12                 & 16          & -            & -          \\
    straighten\_rope                & 0             & 0              & 0         & -         & 0          & -          & -       & -       & 0           & 0           & 0             & 0          & 0                  & 0                  & 0           & -            & -          \\
    setup\_chess                    & 0             & 0              & 0         & 0         & 0          & -          & 0       & -       & 0           & 0           & 0             & 0          & 0                  & 0                  & 0           & -            & -          \\
    scoop\_with\_spatula            & 0             & 0              & 0         & 0         & 0          & 0          & 0       & 0       & 0           & 0           & 0             & 0          & 0                  & 0                  & 0           & 0            & 0         
    \end{tabular}
    \caption{Results for \texttt{MVP} for various perturbations}
    \label{tab:results_mvp}
    \vspace{1cm}
\end{table*}

\begin{table*}[]
    \centering
    \begin{tabular}{lccccccccccccccccc}
    Task Name                       & \myrotcell{No variations} & \myrotcell{All variations} & \myrotcell{MO\_Color} & \myrotcell{RO\_Color} & \myrotcell{MO\_Texture} & \myrotcell{RO\_Texture} & \myrotcell{MO\_Size} & \myrotcell{RO\_Size} & \myrotcell{Light-color} & \myrotcell{Table-Color} & \myrotcell{Table-texture} & \myrotcell{Distractor} & \myrotcell{Background-texture} & \myrotcell{RLBench variations} & \myrotcell{Camera pose} & \myrotcell{Object Friction} & \myrotcell{Object Mass} \\ \hline
    basketball\_in\_hoop            & 84            & 4              & 92        & 4         & 68         & -          & 80      & 84      & 32          & 28          & 88            & 16         & 88                 & 100               & 68         & -            & -        \\
    close\_box                      & 80            & 36             & 8         & -         & -          & -          & 84      & -       & 96          & 56          & 80            & 80         & 84                 & 92                & 68         & -            & -        \\
    close\_laptop\_lid              & 52            & 24             & 80        & -         & -          & -          & 24      & -       & 36          & 48          & 64            & 20         & 68                 & 68                & 56         & -            & -        \\
    empty\_dishwasher               & 0             & 4              & 0         & 0         & -          & 0          & 0       & 0       & 0           & 0           & 0             & 0          & 0                  & 0                 & 0          & -            & -        \\
    get\_ice\_from\_fridge          & 80            & 0              & 68        & 44        & 84         & -          & 56      & 88      & 72          & 60          & 68            & 40         & 84                 & 68                & 80         & -            & -        \\
    hockey                          & 4             & 0              & 0         & 0         & -          & 0          & 0       & 4       & 28          & 36          & 0             & 0          & 0                  & 0                 & 0          & 0            & 0        \\
    meat\_on\_grill                 & 12            & 40             & 16        & 56        & -          & -          & 8       & -       & 28          & 12          & 4             & 12         & 8                  & 76                & 4          & -            & -        \\
    move\_hanger                    & 80            & 0              & 0         & 96        & -          & -          & -       & -       & 0           & 0           & 100           & 8          & 84                 & 84                & 0          & -            & -        \\
    wipe\_desk                      & 0             & 0              & 0         & 0         & 0          & -          & 0       & -       & 0           & 0           & 0             & 0          & 0                  & 0                 & 0          & 0            & 0        \\
    open\_drawer                    & 64            & 0              & 0         & -         & -          & -          & 72      & -       & 68          & 64          & 68            & 52         & 52                 & 72                & 72         & -            & -        \\
    slide\_block\_to\_target        & 0             & 0              & 0         & -         & 0          & -          & 0       & -       & 0           & 0           & 0             & 0          & 0                  & 72                & 0          & -            & -        \\
    reach\_and\_drag                & 84            & 0              & 24        & 52        & 88         & 88         & 92      & 0       & 72          & 52          & 88            & 4          & 88                 & 76                & 80         & 88           & 88       \\
    put\_money\_in\_safe            & 44            & 0              & 68        & 0         & 44         & 28         & 52      & -       & 16          & 16          & 16            & 32         & 36                 & 60                & 64         & -            & -        \\
    place\_wine\_at\_rack\_location & 60            & 12             & 72        & 40        & -          & 72         & 36      & 64      & 88          & 88          & 60            & 32         & 52                 & 56                & 72         & -            & -        \\
    insert\_onto\_square\_peg       & 4             & 0              & 0         & 16        & -          & 12         & 24      & 4       & 8           & 16          & 20            & 4          & 4                  & 8                 & 8          & -            & -        \\
    stack\_cups                     & 0             & 0              & 12        & -         & 12         & -          & 0       & -       & 40          & 12          & 24            & 0          & 16                 & 24                & 20         & -            & -        \\
    turn\_oven\_on                  & 88            & 8              & 40        & -         & -          & -          & 28      & -       & 52          & 36          & 80            & 72         & 92                 & 80                & 80         & -            & -        \\
    straighten\_rope                & 32            & 0              & 20        & -         & 48         & -          & -       & -       & 0           & 28          & 52            & 4          & 92                 & 40                & 68         & -            & -        \\
    setup\_chess                    & 24            & 0              & 4         & -         & 24         & -          & 16      & -       & 0           & 4           & 8             & 0          & 4                  & 4                 & 20         & -            & -        \\
    scoop\_with\_spatula            & 80            & 0              & 16        & 68        & 80         & 88         & 64      & 80      & 44          & 44          & 84            & 0          & 76                 & 88                & 84         & 72           & 68       
    \end{tabular}
    \caption{Results for \texttt{RVT} for various perturbations}
    \label{tab:results_rvt}
    \vspace{1cm}
\end{table*}

\begin{table*}[]
    \centering
    \begin{tabular}{lccccccccccccccccc}
    Task Name                       & \myrotcell{No variations} & \myrotcell{All variations} & \myrotcell{MO\_Color} & \myrotcell{RO\_Color} & \myrotcell{MO\_Texture} & \myrotcell{RO\_Texture} & \myrotcell{MO\_Size} & \myrotcell{RO\_Size} & \myrotcell{Light-color} & \myrotcell{Table-Color} & \myrotcell{Table-texture} & \myrotcell{Distractor} & \myrotcell{Background-texture} & \myrotcell{RLBench variations} & \myrotcell{Camera pose} & \myrotcell{Object Friction} & \myrotcell{Object Mass} \\ \hline
    basketball\_in\_hoop            & 32            & 40             & 56        & 48        & 32         & -          & 60      & 52      & 40          & 40          & 60            & 61         & 44                 & 44                & 56             & -            & -                 \\
    close\_box                      & 0             & 0              & 0         & -         & -          & -          & 0       & -       & 0           & 0           & 0             & 0          & 0                  & 0                 & 0              & -            & -                 \\
    close\_laptop\_lid              & 0             & 0              & 0         & -         & -          & -          & 0       & -       & 0           & 0           & 0             & 0          & 0                  & 0                 & 0              & -            & -                 \\
    empty\_dishwasher               & 0             & 0              & 0         & 0         & -          & 0          & 0       & 0       & 0           & 0           & 0             & 0          & 0                  & 0                 & 0              & -            & -                 \\
    get\_ice\_from\_fridge          & 0             & 0              & 0         & 0         & 0          & -          & 0       & 0       & 0           & 0           & 0             & 0          & 0                  & 0                 & 0              & -            & -                 \\
    hockey                          & 0             & 0              & 0         & 0         & -          & 0          & 0       & 0       & 0           & 0           & 0             & 0          & 0                  & 0                 & 0              & 0            & 0                 \\
    meat\_on\_grill                 & 0             & 0              & 0         & 0         & -          & -          & 0       & -       & 0           & 0           & 0             & 0          & 0                  & 0                 & 0              & -            & -                 \\
    move\_hanger                    & 0             & 0              & 0         & 0         & -          & -          & -       & -       & 0           & 0           & 0             & 0          & 0                  & 0                 & 0              & -            & -                 \\
    wipe\_desk                      & 0             & 0              & 0         & 0         & 0          & -          & 0       & -       & 0           & 0           & 0             & 0          & 0                  & 0                 & 0              & 0            & 0                 \\
    open\_drawer                    & 0             & 0              & 0         & -         & -          & -          & 0       & -       & 0           & 0           & 0             & 0          & 0                  & 0                 & 0              & -            & -                 \\
    slide\_block\_to\_target        & 76            & 80             & 72        & -         & 70         & -          & -       & -       & 60          & 64          & 84            & 76         & 88                 & 80                & 68             & -            & -                 \\
    reach\_and\_drag                & 0             & 0              & 0         & 0         & 0          & 0          & 0       & 0       & 0           & 0           & 0             & 0          & 0                  & 0                 & 0              & 0            & 0                 \\
    put\_money\_in\_safe            & 0             & 0              & 0         & 0         & 0          & 0          & 0       & -       & 0           & 0           & 0             & 0          & 0                  & 0                 & 0              & -            & -                 \\
    place\_wine\_at\_rack\_location & 0             & 0              & 0         & 0         & -          & 0          & 0       & 0       & 0           & 0           & 0             & 0          & 0                  & 0                 & 0              & -            & -                 \\
    insert\_onto\_square\_peg       & 0             & 0              & 0         & 0         & -          & 0          & 0       & 0       & 0           & 0           & 0             & 0          & 0                  & 0                 & 0              & -            & -                 \\
    stack\_cups                     & 0             & 0              & 0         & -         & 0          & -          & 0       & -       & 0           & 0           & 0             & 0          & 0                  & 0                 & 0              & -            & -                 \\
    turn\_oven\_on                  & 0             & 0              & 0         & -         & -          & -          & 0       & -       & 0           & 0           & 0             & 0          & 0                  & 0                 & 0              & -            & -                 \\
    straighten\_rope                & 0             & 0              & 0         & -         & 0          & -          & -       & -       & 0           & 0           & 0             & 0          & 0                  & 0                 & 0              & -            & -                 \\
    setup\_chess                    & 0             & 0              & 0         & -         & 0          & -          & 0       & -       & 0           & 0           & 0             & 0          & 0                  & 0                 & 0              & -            & -                 \\
    scoop\_with\_spatula            & 0             & 0              & 0         & 0         & 0          & 0          & 0       & 0       & 0           & 0           & 0             & 0          & 0                  & 0                 & 0              & 0            & 0               
    \end{tabular}
    \caption{Results for \texttt{Voxposer} for various perturbations}
    \label{tab:results_voxposer}
    \vspace{1cm}
\end{table*}

\end{document}